\documentclass[a4paper]{article}
\usepackage{iwslt18,amssymb,amsmath,epsfig}

\def\reg{{\rm\ooalign{\hfil
     \raise.07ex\hbox{\scriptsize R}\hfil\crcr\mathhexbox20D}}}

\usepackage{graphicx}
\usepackage{url}
\usepackage{soul}
\usepackage{latexsym}
\usepackage{amsmath}
\usepackage{amsxtra}
\usepackage{amssymb}
\usepackage{pifont}
\usepackage{mathrsfs}
\usepackage{array}
\usepackage{ntheorem}
\usepackage{xspace}
\usepackage{numprint}
\usepackage{comment}
\usepackage{setspace}
\usepackage{caption}
\usepackage{subcaption}
\usepackage[printonlyused,nohyperlinks]{acronym}
\usepackage{pifont}
\usepackage{enumitem}

\usepackage{tabularx}
\usepackage{booktabs}
\usepackage{rotating}
\usepackage{multirow}

\usepackage{todonotes}

\usepackage{xcolor}

\name{}
 \makeatletter
 \def\name#1{\gdef\@name{#1\\}}
 \makeatother
 \name{{\em Rohit Gupta$^{\star}$\thanks{$^{\star}$\ {\tt rohitg1594@gmail.com} Work done while at Naver Labs Europe.}, 
 Laurent Besacier$^{\dagger \mathsection}$\thanks{$^{\mathsection}$\ {\tt laurent.besacier@univ-grenoble-alpes.fr}}, 
 Marc Dymetman$^{\dagger}$\thanks{$^{\dagger}$\ {\tt {\{marc.dymetman,matthias.galle\}@naverlabs.com}}}, 
 Matthias Gall\'e$^{\dagger}$}}
\address{$^{\dagger}$Naver Labs Europe, France \\
$^{\mathsection}$LIG - Universit\'e Grenoble Alpes, France}
\title{Character-based NMT with Transformer}
\begin{document}
\maketitle
\begin{abstract}
    Character-based translation has several appealing advantages, but its performance is in general worse than a carefully tuned BPE baseline.
    In this paper we study the impact of character-based input and output with the Transformer architecture.
    In particular, our experiments on EN-DE show that character-based Transformer models are more robust than their BPE counterpart, both when translating noisy text, and when translating text from a different domain.
    To obtain comparable BLEU scores in clean, in-domain data and close the gap with BPE-based models we use known techniques to train deeper Transformer models.
    
\end{abstract}

\section{Introduction}

Character-level NMT models have some compelling characteristics.
They do not suffer from out-of-vocabulary problem and avoid tedious and language-specific pre-processing  that adds yet another hyper-parameter to tune.
In addition, they have been reported to be more robust when translating noisy text~\cite{WhatIsATranslationUnit} and -- when using the same architecture -- are more compact to store.
Those two characteristics are particularly important for translating user-generated content or spoken language, which is often noisy due to transcription errors.
On the drawbacks, they tend to perform worse in translation quality than using words or Byte-Pair Encoding (BPE)~\cite{BPEemb,QuasiRNN} segmentations.
In this paper we perform extensive experiments to measure the possibilities of character-based Transformer models in such scenarios.

For LSTM-based architectures, \cite{RevisitingCharacter} showed that it was possible to obtain similar performance at the cost of training deeper networks.
The current state-of-the-art model for NMT (and NLP in general) however is Transformer  \cite{Vaswani2017} and to our knowledge no equivalent study has been reported for that architecture.
In this paper, we analyze the impact of character-level Transformer models versus BPE-based  and evaluate them on four axes: 
\begin{itemize}
    \item translating on clean vs noisy text, 
    \item in-domain vs out-of-domain conditions, 
    \item training in low and high-resource conditions, 
    \item impact of different network depths.
\end{itemize}

Our experiments are for EN-DE, on news (WMT) and TED talks (IWSLT).
The results show that:
\begin{itemize}
    \item it is possible to narrow the gap between BPE and character-level models with deeper encoders
    \item character-level models are more robust to lexicographical noise than BPE models out of the box
    \item character-level models cope better with test data that is far apart from the training data set
\end{itemize}

\section{Related Work}

\textbf{Input representations} 
For deciding what should be the atomic input symbols, the most intuitive way seems to be to use words as tokens.
Continous word representation~\cite{Word2Vec,Glove} have shown tremendous impact in NLP application~\cite{PosTagging,Yamada,QuestionAnswering,SNLI}.
While some of those representations exploit morphological features~\cite{FastText}, such representation still face challenges due to the large vocabulary needed and to out of vocabulary words. To circumvent these issues, some works learn language representations directly at the character level and disregard any notion of word segmentation. This approach is  attractive due to its simplicity and ability to adapt to different languages. It has been used in a wide range of NLP tasks such as language modeling \cite{Mikolov2011SUBWORDLM,AlRfou}, question answering~\cite{CharacterQA} and parsing~\cite{CharacterParsing}.

For translation, character level models in NMT initially showed unsatisfactory performance \cite{VilarNMT,NeubigCharacterNMT}. The two earliest models with positive results were \cite{CharacterNMT} and \cite{CostaCharacterNMT}. They compose word representations from their constituent characters and as such require an offline segmentation step to be performed beforehand. \cite{FullyCharacterNMT} {was} able to obviate this step by composing representations of ``pseudo'' words from characters using convolutional filters and highway layers \cite{HighwayLayers}. The {previous} methods introduce special modifications to the NMT architecture in order to work at the character level. \cite{RevisitingCharacter}, on the other hand, {uses} a vanilla (LSTM based) NMT system to achieve superior results at the character level. In a different direction, \cite{LearningToSementFavours} {proposes} to dynamically learn segmentation  informed by the NMT objective. The authors discovered that their model prefers to operate on (almost) character level, providing support for purely character-based NMT.
%from a novel angle

A common approach for dealing with the open vocabulary  issue  is  to  break  up  rare  words  into sub-word  units \cite{SennrichBPE,GoogleNMT}. 
BPE~\cite{SennrichBPE} is the standard technique in {NMT} and has been applied to great success to many systems \cite{StrongerBaselines,WS2017}. {BPE} has one hyperparameter: number of merge operations $n$. The optimal $n$ depends on many factors including {NMT} architecture, language characteristics and size of training dataset. \cite{StrongerBaselines} explored several hyperparameter settings, including number of {BPE} merge operations, to establish strong baselines for NMT in LSTM-based architectures.  They recommended \textit{``32K as a generally effective vocabulary size and 16K as a contrastive condition when building systems on less than 1 million parallel sentences''}. \cite{CallForPrudence} does a thorough study on the impact of $n$ for both {LSTM} and Transformer architectures. The authors conclude that there is in fact no optimal $n$ for {LSTM}; it can be very different depending on the dataset and language pair. However, for the Transformer, the best {BPE} size is between character level and 10k.
 
\textbf{Deep models} Until recently it was very hard to train very deep models with the standard Transformer architecture. The training dynamics tended to be unstable with degradation of performance for deeper models. \cite{BengioUnderstanding} argued that the main culprit was the then in vogue non-linearity function: sigmoid. It saturates for deep models, blocking gradient information from flowing backward. Consequently, \cite{RELU} proposed the ReLU non-linearity, which is the \emph{de facto} standard today. Though this simple technique allows one to train deeper models than before, it is not sufficient for very deep models with more than 30 layers. Residual connections \cite{ResNET} were formulated so that the consequent layers have direct access to the layer inputs in addition to the usual forward functions. This simple tweak makes it possible to train models of up to \numprint{1000} layers, achieving {SOTA} on an image classification benchmark. \cite{TransparentAttention} find it hard to train the Transformer with more than 10 encoder layers. It proposes \textit{transparent attention}, wherein encoder-decoder attention is computed over a linear combination of the outputs of all encoder layers. This alleviates the gradient vanishing or exploding problem and is sufficient to train Transformer with an encoder of 24 layers. \cite{DeepTransformer} extends \cite{TransparentAttention} and achieves slight but robust improvements. 
 
\textbf{Robustness} Machine learning systems can be brittle. Small changes to the input can lead to dramatic failures of deep learning models \cite{IntriguingProperties,ExplainingAndHarnessing}. For {NMT}, \cite{BelinkovNoise} studied robustness to lexicographical errors. They found both character and {BPE} to be very sensitive to such errors with severe degradation in performance (out-of-the-box robustness of character models was however slightly better than the one of BPE models). They proposed two techniques to improve robustness of {NMT} models: structure-invariant word  representations  and training  on  noisy  texts. These techniques are sufficient to make a character based model simultaneously robust to multiple kinds of noise. \cite{TrainingOnSyntheticNoise} and \cite{ImprovingRobustness} also report similar findings, namely that training on a balanced diet of synthetic noise can dramatically improve robustness on synthetic noise. While \cite{ImprovingRobustness} leverage the noise distribution in the test set, \cite{TrainingOnSyntheticNoise} does not.
Dealing with noisy data for {NMT} can also be seen as a domain adaptation problem \cite{SixChallenges}. The main discrepancy is between the distribution of the training data and test data, also known as \textit{domain shift} \cite{DataShiftInML}. Many different approaches have been studied to train with multiple domains: \cite{LiAdaptive} and \cite{Axelrod} include data from the target domain into the training set directly without any modifications, \cite{KobusCS16} introduce domain tags to differentiate between different domain, and finally, \cite{ZhangTopic} and \cite{ChenTopic} use a topic model to add topic information about the domain during training.
\cite{berard2019} describes the winning entry to the WMT'19 robustness challenge.

\section{Representation units for Transformer}

\subsection{Character vs BPE models}
We experimented on two language directions, namely, German-English (DE-EN) and English-German (EN-DE). For DE-EN, we consider two settings: \textit{high resource} and \textit{low resource}. For high resource, we concatenate the commoncrawl \cite{commoncrawl} and Europarl \cite{europarl} corpora. We used the {WMT} 2015 news translation test set as our validation set and WMT 2016 as the test set. For the low resource setting, we used the IWSLT 2014 corpus \cite{IWSLT14}, consisting of transcriptions and translations of TED Talks.\footnote{\url{https://www.ted.com/talks}} We used the official train, valid and test splits. In EN-DE, we used the same setup as the low resource setting of DE-EN in the opposite direction. The {IWSLT}14 dataset is much smaller than the {WMT} corpus used originally by \cite{Vaswani2017}. Therefore, for the low resource setting we use a modified version of the Transformer base architecture with approximately 50M parameters as compared to 65M for Transformer base. For the high resource setting we use the unmodified Transformer base architecture.

The training details for the low resource setting are as follows. Training is done on 4 GPUs with a max
batch size of \numprint{4000} tokens (per GPU). We train for
150 and 60 epochs in the low and high resource settings respectively, while saving a checkpoint after every epoch and average the 3
best checkpoints according to their perplexity on a
validation set. In the low resource setting, we test all 6 combinations of
dropout in $[0.3, 0.4, 0.5]$ and learning rate in
$[5, 10] \times 10^{-4}$. Using the best dropout and learning rate combination, 5 models (with different random seeds) are trained. Whereas for the high resource setting, we tune dropout in $[0.1, 0.2, 0.3]$ and set the max learning rate to be $5 \times 10^{-4}$. Due to the significantly larger computational requirements for this dataset, we only train one model.

The average performance and standard deviation (over 5 models) on test set are shown Table \ref{table:bpe impact}. The following conclusions can be drawn from the numbers:
 
 \begin{table}[]
\centering
    \begin{tabular}{c|ccc}
    Vocab Size & DE-EN (low) & EN-DE (low) & DE-EN (high)\\ \hline
    Char       & 33.7 $\pm$ 0.1 & 26.7 $\pm$ 0.1 & 36.3\\
    \numprint{1000}       & 34.0 $\pm$ 0.2 & 26.8 $\pm$ 0.1 & --\\
    \numprint{2000}       & 34.4 $\pm$ 0.2 & 27.1 $\pm$ 0.1 & --\\
    \numprint{5000}       & \textbf{35.0} $\pm$ 0.0 & 27.4 $\pm$ 0.1 & 36.2\\
    \numprint{10000}      & 34.6 $\pm$ 0.2 & \textbf{27.6} $\pm$ 0.1 & --\\
    \numprint{20000}      & 30.5 $\pm$ 0.2 & 25.3 $\pm$ 0.1 & --\\
    \numprint{30000}      & 28.3 $\pm$ 0.1 & 24.3 $\pm$ 0.1 & \textbf{37.2}\\
    \numprint{40000}      & 27.1 $\pm$ 0.1 & 23.7 $\pm$ 0.2 & --\\
    \numprint{50000}      & 26.2 $\pm$ 0.3 & 23.1 $\pm$ 0.2 & --\\
    \end{tabular}
    \caption{Impact of {BPE} vocab size on {BLEU}.}
    \label{table:bpe impact}
\end{table}

\begin{enumerate}[topsep=0pt, partopsep=0pt]
    \item \textbf{Vocabulary matters for low resource}. The impact of vocabulary size is significant in the low resource setting, {BLEU} scores differ by over 8 points for DE-EN and close to 5 {BLEU} for EN-DE. For the high resource setting, the effect of vocabulary size is minimal over a large range.
    \item \textbf{Optimal {BPE} is small for low resource}. The optimal vocabulary size is either \numprint{5000} for DE-EN or \numprint{10000} for EN-DE. In the high resource setting, 30K is optimal and we corroborate the standard choice.
    \item \textbf{Character level models are competitive}. Though the character-level models are not able to beat the best {BPE} models, they are surprisingly competitive without any modifications to the architecture.
\end{enumerate}

\subsection{Noisy vs clean}
We introduce the following four different types of character level synthetic noise with an associated noise probability $p$. We respect word boundaries by only applying noise within the word.

\begin{enumerate}[topsep=0pt, partopsep=0pt]
    \item \texttt{delete}. Randomly delete a character except for punctuation or space.
    \item \texttt{insert}. Insert a random character.
    \item \texttt{replace}. Replace the current character,
    \item \texttt{switch}. Switch the position of two consecutive characters. We do not apply this for first and last character of a word.
    \item \texttt{all}. With a probability of $p / 4$, introduce one of the noises listed above.
\end{enumerate}

\smallskip

For DE-EN, we also experiment with \texttt{natural} noise. We follow \cite{BelinkovNoise} and use their dataset of naturally occurring noise in German.\footnote{We accessed the dataset from \url{https://github.com/ybisk/charNMT-noise/blob/master/noise/de.natural}.} It combines two projects:  RWSE Wikipedia Revision Dataset \cite{ZeschGermanNoise} and the  MERLIN  corpus  of  language  learners  \cite{MERLIN}. These corpora were created to measure spelling difficulty and consist of word lists, wherein a correct German word has associated with it a list of common mistakes. For example  word ``Familie'' can be replaced in our \texttt{natural} test set by ``Famielie'', ``Fammilie'', etc.

\begin{figure}
    \centering
    \includegraphics[width=.5\textwidth]{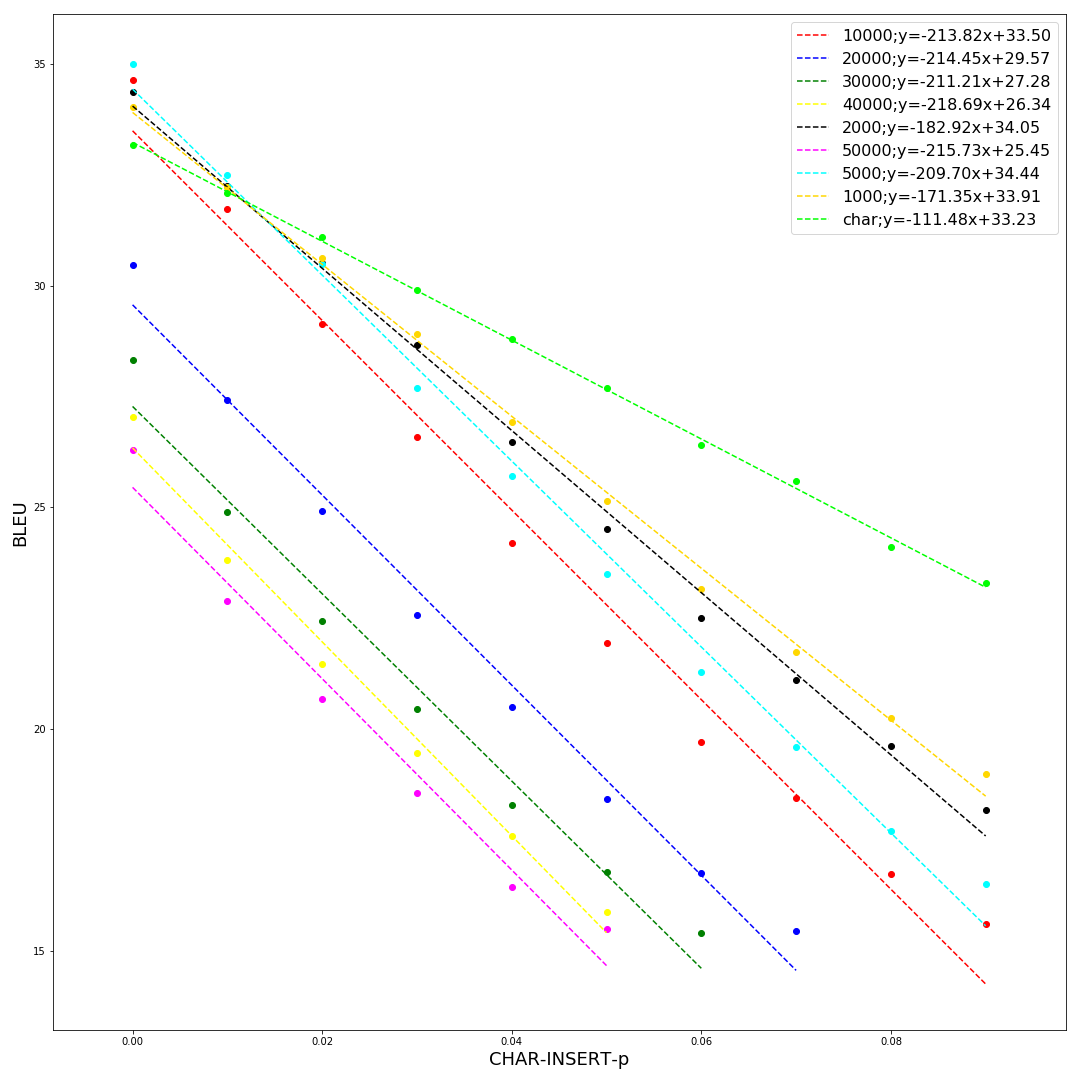}
   \caption{Degradation of translation quality with increasing noise (character insertion). The slope of the curves (sensibility) is smaller and shows character level is more robust here.  }
    \label{fig:robustnessP}
\end{figure}

For each noise type, we create ten different noisy versions of the test set with different noise probabilities. For synthetic noise, noise proportions were $p=1$, $2$, $\dots$, $10\%$, whereas for \texttt{natural} noise, noise proportions were $p=10$, $20$, $\dots$, $100\%$. 
Note that the $100\%$ \texttt{natural} noise test set does not have all its tokens transformed. 
A majority of words have no naturally occurring spelling error in \cite{BelinkovNoise}'s dataset.
We then compute the BLEU test score on that noisy test data, for each $p$ and for models trained with different vocabulary sizes.
A representative such plot can be seen in Fig.~\ref{fig:robustnessP}, for the case of using \textit{insertion} and where each line corresponds to one vocabulary size.

We calculate the BLEU scores on noisy test sets with different noise probabilities and for each data series, we compute a linear regression:  

\begin{equation}
    \text{BLEU} \approx \beta p + \alpha,
\end{equation}

\noindent where $p$ is the noise probability, the slope $\beta$ is the ``sensitivity'' of the NMT system to that type of noise and $\alpha$ is the intercept. Closer to 0 means that the system is more robust to that kind of noise, and a value of $-100$ indicates that for each additional percentage point of noise the system loses 1 BLEU point. 
Those values can be seen in 
%the legend box of Fig.~\ref{fig:robustnessP}.
Fig.~\ref{fig:noise_sensitivities} where we plot the values of $\beta$ vs the vocabulary size.
We conclude from that:

\begin{figure*}[h]
	\centering
    \begin{subfigure}[b]{.4\linewidth}
    \centering
		\includegraphics[width=\linewidth]{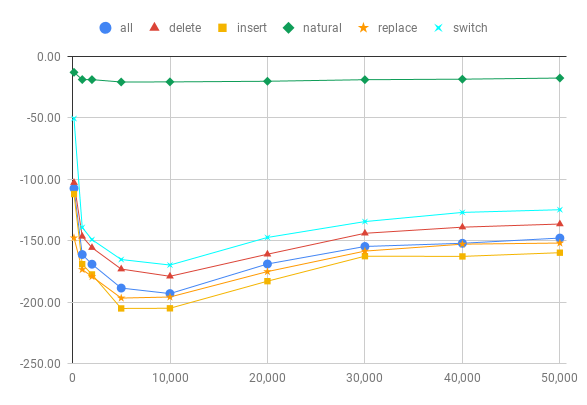}
	    \caption{DE-EN (low)}
	    \label{fig:delete-iwslt14-de-en}
	\end{subfigure}
	\begin{subfigure}[b]{.4\linewidth}
	    \centering
		\includegraphics[width=\linewidth]{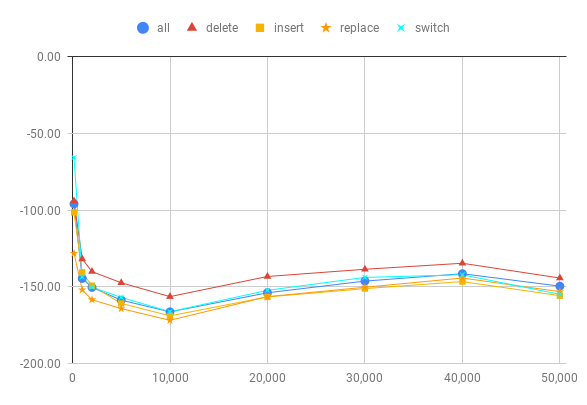}
	    \caption{EN-DE (low)}
	    \label{fig:insert-iwslt14-de-en}
	\end{subfigure}
	
	\begin{subfigure}[b]{.4\linewidth}
	    \centering
		\includegraphics[width=\linewidth]{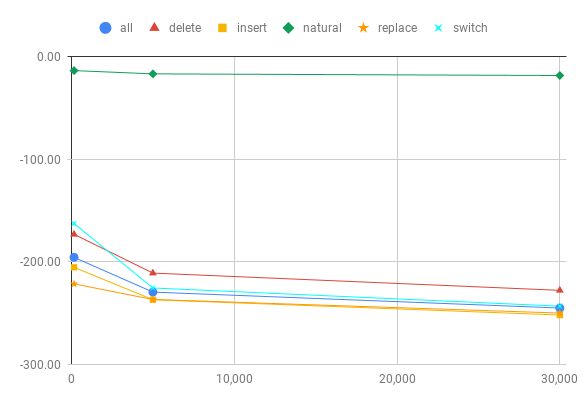}
	    \caption{DE-EN (high)}
	    \label{fig:insert-iwslt14-de-en}
	\end{subfigure}
	\caption{Noise sensitivity vs vocab size for models trained on clean data. Character-level models are shown with zero vocabulary size. Sensitivity values closer to zero mean that the model is more robust to that kind of noise.}
\label{fig:noise_sensitivities}
\end{figure*}

\begin{figure*}
    \centering
    \includegraphics[width=.6\textwidth]{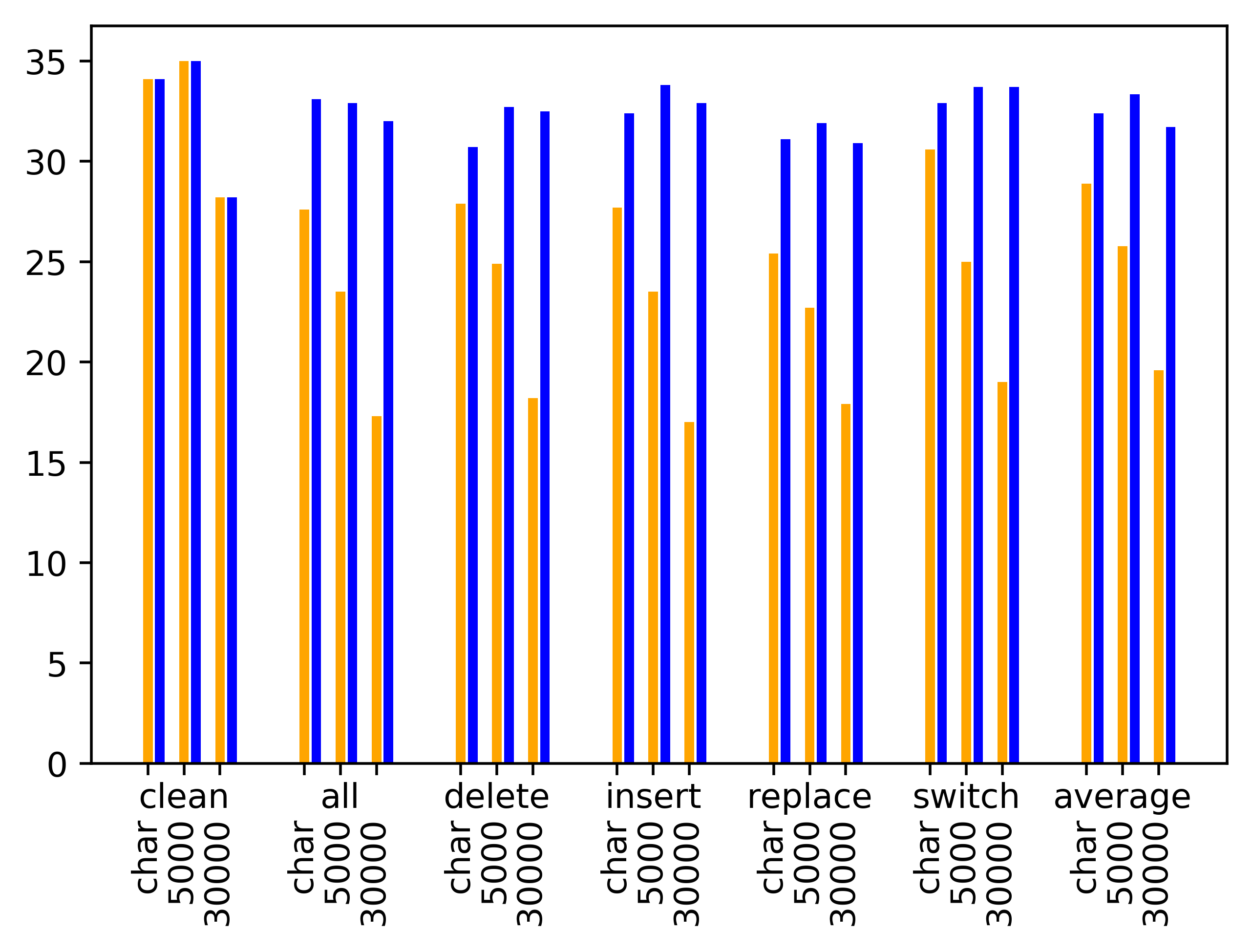}
    \caption{{BLEU} scores for DE-EN models trained and tested on  different noise conditions. The first (orange) column refers to training on clean data and testing on noised data; the second (blue) trained and tested on matched noise (which is the same for the \textit{clean} group)}
    \label{fig:noiseDEEN}
\end{figure*}

\begin{figure*}
    \centering
    \includegraphics[width=.6\textwidth]{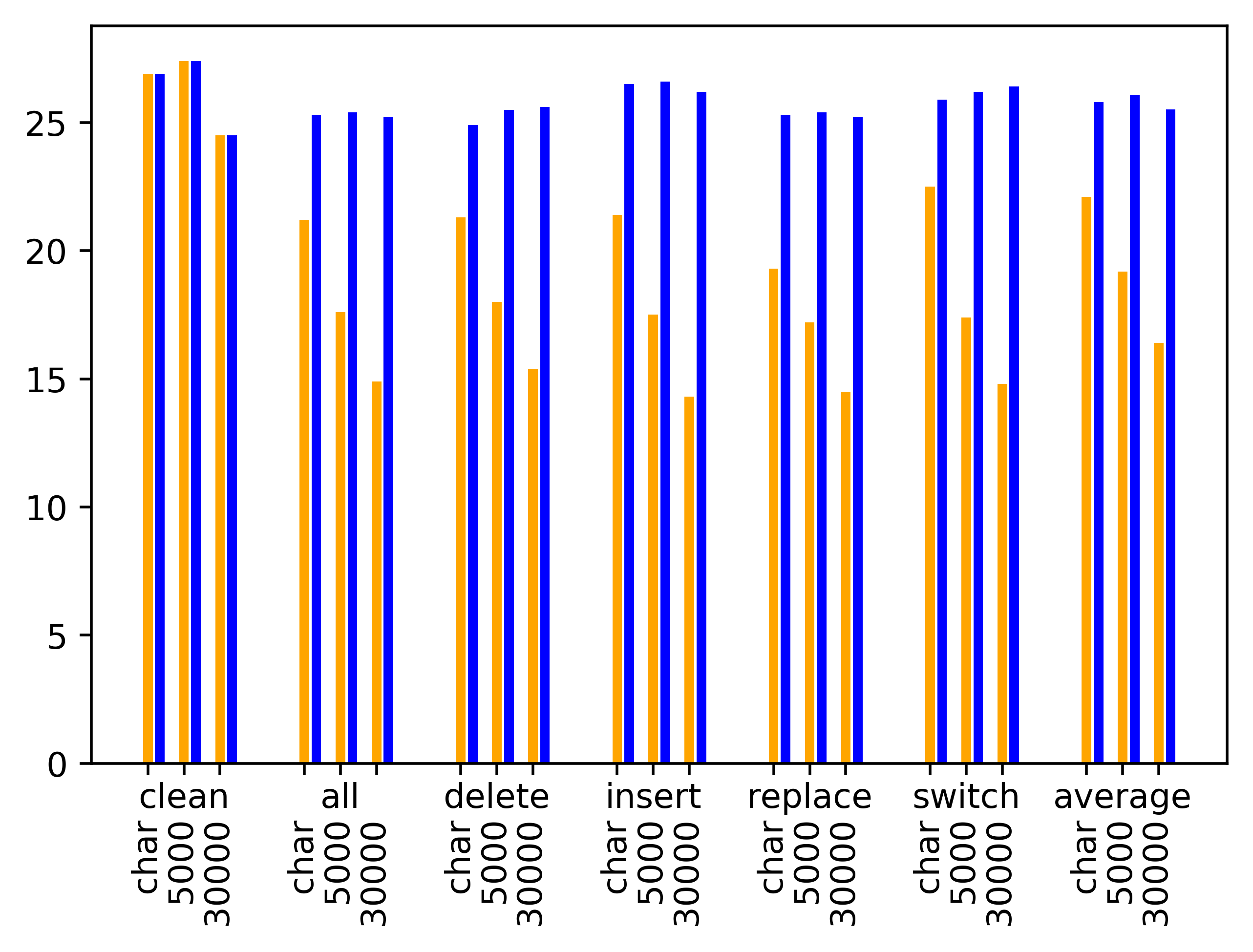}
    \caption{{BLEU} scores for EN-DE models trained and tested on  different noise conditions. The first (orange) column refers to training on clean data and testing on noised data; the second (blue) trained and tested on matched noise (which is the same for the \textit{clean} group)}
    \label{fig:noiseENDE}
\end{figure*}

\begin{enumerate}
    \item \textbf{Degradation with noise}. Out of the box, both BPE and character level models are very sensitive to lexicographic noise. BPE models lose as much as 2 BLEU points for each percentage increase in noise, whereas character level models lose as much as 1.5 BLEU.
    
    % \item \textbf{Behaviour of different noises.} BPE models are roughly equally sensitive to all kinds of synthetic noise. Sensitivity to \texttt{natural} noise is smaller by a factor of ten. \texttt{natural} noise is structured and predictable and the model is able to correct for them up to a large extent. On the other hand, our synthetic noise models are much less structured, making it hard for the models to adapt.
    % Character level models are more sensitive to certain kinds of noises than other. They are relatively very robust to \texttt{switch}, approximately equally robust to \texttt{delete} and \texttt{insert} and least robust to \texttt{replace}. We hypothesize that this could be due to \texttt{switch} only changing the positional encodings locally. The content embeddings remain intact. In contrast, \texttt{replace} preserves positional encodings but changes the content embeddings.
    
    % Correction attempt by MD (17.10.2019) after reading reviewers' comments
    \item \textbf{Behaviour of different noises.} BPE models are roughly equally sensitive to all kinds of synthetic noise. 
    Character level models are more sensitive to certain kinds of noises than other. They are relatively very robust to \texttt{switch}, approximately equally robust to \texttt{delete} and \texttt{insert} and least robust to \texttt{replace}. We hypothesize that this could be due to \texttt{switch} only changing the positional encodings locally. The content embeddings remain intact. In contrast, \texttt{replace} preserves positional encodings but changes the content embeddings.
    Sensitivity to \texttt{natural} noise is much smaller than to synthetic noise overall, probably due to the fact that increasing the noise level does not have any effect on words that are not listed in the \cite{BelinkovNoise} dataset. %MD: this is my wild guess, which might be totally wrong ...

    \item \textbf{Character level models are more robust}. For each kind of noise, character level models are less sensitive than all of the BPE models. They are particularly robust to \texttt{switch}, where they are more than twice as robust as the best performing BPE models. Though character level models start out at a worse footing than the best BPE models, after applying only 1-2 \% (for synthetic noise) of noise in the test set, character level models perform better.
\end{enumerate}

We also experiment with a simple method to robustify models to the noise. We introduce the aforementioned noises into the training data as well. Thereafter, we test on both clean and noisy data. In consideration for time and computational resources, we choose two representational BPE vocabularies -- \numprint{5000} for small vocab size and \numprint{30000} for large vocab size. We also train character level models with noisy data. We only consider synthetic character level noises with noisy probability set to $5\%$ in the low resource setting. 
The results are shown in Fig.~\ref{fig:noiseDEEN} and \ref{fig:noiseENDE} for DE-EN and EN-DE respectively. Each group of two represents training on clean/matched noise; for 3 different vocabulary sizes and 6 different types of noise.
The full results of these sets of experiments are in Appendix \ref{sect:robustnessNoise_full}.
The following conclusions are apparent.

\begin{enumerate}
    \item \textbf{Adding noise helps}. Training on similar type of noisy data  improves performance for all vocabularies.\footnote{We also observed that certain kinds of noise also improve robustness for other noises (results  with unmatched train and test conditions not reported here).}
    
    \item \textbf{BPE is as robust as character}. By training on similar kinds of noise in the training data, we are able to robustify BPE models to the same level as character level models without sacrificing too much performance on the clean test set.
    
    \item \textbf{Effect on clean data}. We observed (see Tables \ref{tab:train_noisy_data_de_en_full} and \ref{tab:train_noisy_data_en_de_full} in the Appendix) that for small vocabularies (character-level and BPE \numprint{5000}), training with noise in training data had a small detrimental effect when testing on clean data. However, in the case of BPE \numprint{30000}, training on noisy data significantly boosted performance (eg, improvement of 6 BLEU for DE-EN and 1.7  for EN-DE when training with \texttt{delete} and testing on clean data). We hypothesize that the increased diversity of tokens during training (due to the presence of noise) acts as a regularizer boosting performance on the test set.
    
\end{enumerate} 

\subsection{In-domain vs out-of-domain}
We test the low and high resource models on the following in and out of domain datasets:

\begin{enumerate}[topsep=0pt, partopsep=0pt]
    \item \textbf{newstest 2016}. News text from WMT 2016 news translation task
    \item \textbf{WMT Biomedical}. Medline abstracts from WMT 2018 biomedical translation task.
    \item \textbf{WMT-IT}. Hardware and software troubleshooting answers from WMT 2016 IT domain translation task.
    \item \textbf{Europarl}. The first \numprint{3000} sentences from Europarl corpus \cite{europarl}. Proceedings of the European Parliament.
    \item \textbf{commoncrawl}. The first \numprint{3000} sentences from commoncrawl parallel text corpus \cite{commoncrawl}.
\end{enumerate}

\begin{table*}[!htb]
\centering
\begin{tabular}{c|c|cc|cc|cc}
              &          & \multicolumn{2}{c|}{DE-EN} & \multicolumn{2}{c|}{EN-DE} & \multicolumn{2}{c}{DE-EN (high)} \\
Dataset       & \# Sents & \% Unseen & PPL            & \% Unseen & PPL            & \% Unseen & PPL  \\          
\hline
IWSLT14       & \numprint{6750}    & 4.4       & 583            & 2.2       & 282            & 2.0       & 740   \\
WMT-IT        & \numprint{2000}    & 14.4      & \numprint{2540}          & 13.2      & \numprint{2322}          & 5.8       & 996    \\
WMT-Bio.      & 321      & 20.0      & \numprint{5540}          & 12.1      & \numprint{3035}          & 9.2       & \numprint{3404}    \\
newstest 2016 & \numprint{2999}    & 12.7      & 2,712          & 9.0       & 1,659          & 4.5       & \numprint{1703}    \\
Europarl      & \numprint{3000}    & 9.0       & 1,765          & 4.4       & 771            & 0         & 10    \\
commoncrawl   & \numprint{3000}    & 17.8      & \numprint{5024}          & 12.6      & \numprint{2711}          & 0         & 9    \\ 
\hline
avg           & \numprint{3011}    & 13.0      & \numprint{3022}          & 8.9       & 1,797         & 3.6       & \numprint{1144}    \\
\end{tabular}
\caption{Similarity metrics between test sets and  training sets.}
\label{tab:similarity_data}
\end{table*}

%\begin{tabular}{c|c|cc|cc|cc}
%              &          & \multicolumn{2}{l|}{\hspace{1.5cm} DE-EN}  & %\multicolumn{2}{l}{\hspace{1.5cm} EN-DE} & \multicolumn{2}{l}{\hspace{1.5cm} %DE-EN (high)}   \\
%Dataset       & \# Sents & Avg. Seen & \% Unseen & Avg. Seen & \% Unseen & %Avg. Seen & \% Unseen  \\ \hline
%IWSLT14       & \numprint{6750}    & \numprint{11009}    & 4.4       & %\numprint{19935}    & 2.2       & \numprint{371434}   & 2.0        \\
%MT-IT        & \numprint{2000}    & \numprint{12354}    & 14.4      & %\numprint{23372}    & 13.2      & 476,483   & 5.8        \\
%MT-Bio.      & 321      & \numprint{10161}    & 20.0      & \numprint{18694} %  & 12.1      & \numprint{465074}   & 9.2        \\
%ewstest 2016 & \numprint{2999}    & \numprint{9614}     & 12.7      & %numprint{18705}    & 9.0       & \numprint{397814}   & 4.5        \\
%Europarl      & \numprint{3000}    & \numprint{11626}    & 9.0       & %\numprint{23448}    & 4.4       & \numprint{499919}   & 0\\
%commoncrawl   & \numprint{3000}    & \numprint{9134}     & 17.8      & %\numprint{19027}    & 12.6      & \numprint{381046}   & 0\\ \hline
%avg           & \numprint{3011}    & \numprint{10650}    & 13.0      & %\numprint{20530}    & 8.9       & \numprint{431962}   & 3.6
%\end{tabular}

%\caption{Similarity metrics between test sets and  training sets.}
%\label{tab:similarity_data}
%\end{table*}

We provide two similarity metrics between the training and test sets in Tables \ref{tab:similarity_data}.  
%``Avg. Seen'' is the average number of times a word in the test set is seen in the training corpus. 
``\% Unseen'' is the percentage of words in the test set that are not present in the training corpus. 
``PPL'' is the perplexity measure of the test set using a language model trained on the training data. 
We used the \textit{kenlm}\footnote{\url{https://github.com/kpu/kenlm}}  toolkit with Kneser-Ney smoothing~\cite{Kneser-Ney} and context size of 4.

\begin{figure*}[h]
	\centering
	\begin{subfigure}[b]{.33\linewidth}
	    \centering
		\includegraphics[width=\linewidth]{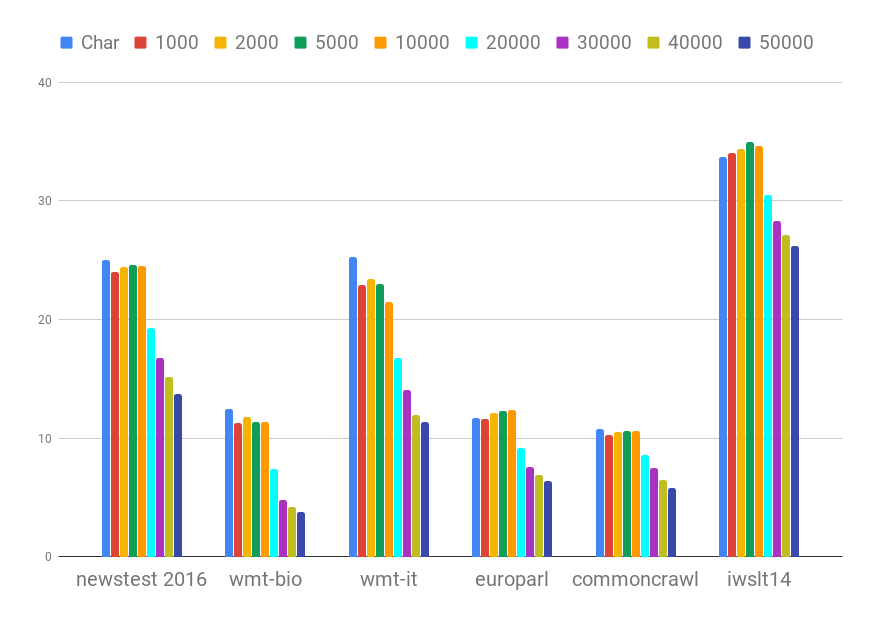}
	    % \caption{EN-DE (low)}
	    \caption{DE-EN (low)}
	    \label{fig:insert-iwslt14-de-en}
	\end{subfigure}
    \begin{subfigure}[b]{.33\linewidth}
    \centering
		\includegraphics[width=\linewidth]{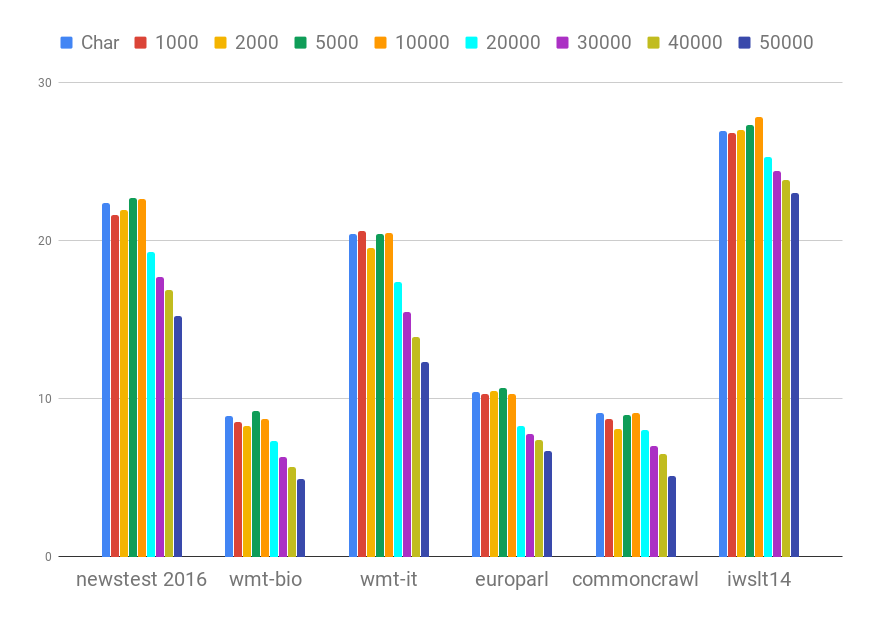}
	    % \caption{DE-EN (low)}
	    \caption{EN-DE (low)}
	    \label{fig:delete-iwslt14-de-en}
	\end{subfigure}
	\begin{subfigure}[b]{.33\linewidth}
	    \centering
		\includegraphics[width=\linewidth]{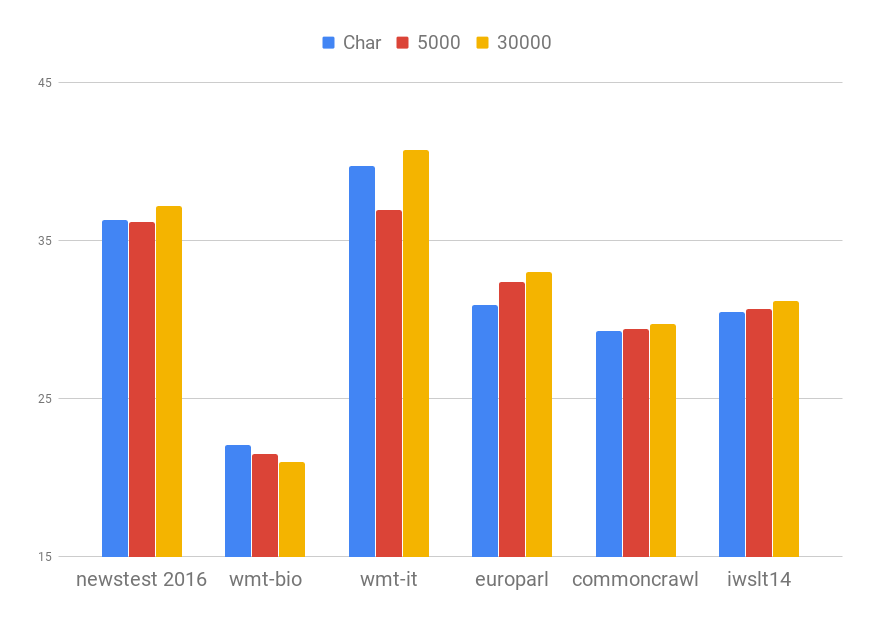}
	    \caption{DE-EN (high)}
	    \label{fig:insert-iwslt14-de-en}
	\end{subfigure}
	\caption{BLEU scores for different vocabularies on test sets from different domains.}
\label{fig:domain_shift}
\end{figure*}

We show results in Figure \ref{fig:domain_shift} and conclude the following:

\begin{enumerate}
    \item \textbf{DE-EN low resource}. Character level models are better for all out of domain datasets, except for Europarl. Recall from Table~\ref{tab:similarity_data} that Europarl also has the least proportion of unseen words. This suggests that character level models outperform {BPE} when evaluated on data sufficiently different from the training domain in this low resource setting.
    \item \textbf{DE-EN high resource}. Character level models are now only better when testing on the WMT-Biomedical test set. We see from Table~\ref{tab:similarity_data} that it also has the largest proportion of unseen words. For all other test sets, {BPE} \numprint{30000} leads to the best {BLEU} scores. 
    \item \textbf{EN-DE low resource}. We see similar performance for in and out of domain data. Good {BPE} models on in-domain test set are still better on out-of-domain test sets. A possible explanation is the lower proportion of unseen words as compared to German and seeing the words more frequently in the training corpus.
\end{enumerate}

\subsection{Deeper character-based Transformers}

For other architectures, training deeper models had a very positive impact on character-based translations~\cite{RevisitingCharacter}, however similar studies have not been reported using the Transformer. Due to computational constraints, we experiment only on DE-EN language pair in the low and high resource settings and train only one model for each configuration.

\begin{figure*}
    \centering
    \includegraphics[width=.6\textwidth]{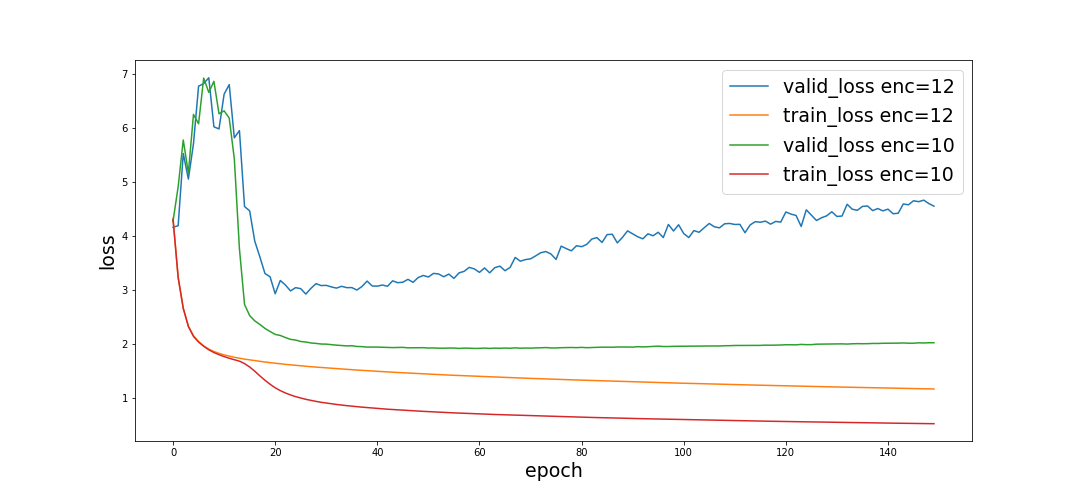}
    \caption{Degradation observed with the standard Transformer architecture when going from 10 to 12 layers.}
    \label{fig:catastrophic}
\end{figure*}

\subsubsection{Low resource}

We train models from 6 to 16 encoder layers for character level and {BPE} \numprint{5000}, the best performing vocabulary size in our preliminary experiments.
We fix learning rate to $5 \times 10^{-4}$ and tune dropout in $[0.3, 0.4, 0.5]$.
First, we do not perform any modifications to the Transformer architecture.
In particular, this means that layer normalization takes place \textit{after} each sub layer.  To train deeper models, following \cite{DeepTransformer}, we place the layer normalization step \textit{before} each layer, and also experiment with transparent attention~\cite{TransparentAttention}. 

In contrast to \cite{RevisitingCharacter}, we see a degradation of performance with increasing depth for post-normalization (Figure  \ref{fig:catastrophic} illustrates this with the standard Transformer architecture when going from 10 to 12 layers), but the simple trick of switching the sequence of performing layer normalization is sufficient to train models with up to 32 layers in the encoder. 
We therefore report only results using pre-normalization in Table \ref{tab:deep_low}. 
While adding transparent attention is beneficial for almost all depths in the character level models, it gives mixed results for the {BPE} \numprint{5000} model. Hence, we see that by training deeper models we are able to marginally improve performance for both vocabularies. For character level models we improve by 1 {BLEU} points (from 33.7 to 34.7). For \numprint{5000}, the gain is a more modest 0.4 {BLEU} points (from 35 to 35.4). We are able to narrow but not close the gap between character and {BPE}.

%\small\addtolength{\tabcolsep}{-3pt}
%% Results with Post-Norm
\begin{table}[!htb]
\centering
%\begin{tabular}{c|ccc|ccc}
  %  & \multicolumn{3}{c|}{Char}                     & \multicolumn{3}{c}{5,000}                            \\
%enc & Post Norm & Pre Norm & Pre Norm + Trans. & Post Norm   & Pre Norm      & Pre Norm + Trans. \\ \hline
%6   & 33.7      & 33.4     & 33.2                   & \textbf{35} & 34.6          & 34.6                   \\
%12  & -         & 33.8     & 34.5                   & 33.5        & \textbf{34.8}         & \textbf{34.8}                 \\
%16  & -         & 33.5     & 34.5                   & 31.2           & \textbf{35.2} & 34.9                   \\
%20  & -         & 34.4     & {\ul 34.7}                 & -           & \textbf{35.3} & 34.9                   \\
%24  & -         & 34.1     & {\ul 34.7}                  & -           & 35.1          & {\ul \textbf{35.4}}    \\
%28  & -         & 34.4     & 34.3                   & -           & \textbf{35}   & \textbf{35}            \\
%32  & -         & 34.1     & 34.5                   & -           & 34.7          & \textbf{35.2}         
%\end{tabular}
\begin{tabular}{c|cc|cc}    & \multicolumn{2}{c|}{Char}     & \multicolumn{2}{c}{\numprint{5000}}    \\
enc                         &  PreN      & PreN+T           & PreN              &     PreN+T         \\ \hline
6                           &  33.4      & 33.2             & \textbf{34.6}     &     \textbf{34.6}                   \\
12                          &  33.8      & 34.5             & \textbf{34.8}     &     \textbf{34.8}                 \\
16                          &  33.5      & 34.5             & \textbf{35.2}     &     34.9                   \\
20                          &  34.4      & 34.7             & \textbf{35.3}     &     34.9                   \\
24                          &  34.1      & 34.7             & 35.1              &     \textbf{35.4}    \\
28                          &  34.4      & 34.3             & \textbf{35}       &     \textbf{35}            \\
32                          &  34.1      & 34.5             & 34.7              &     \textbf{35.2}         
\end{tabular}

\caption{Results for the low resource setting. ``PreN + T'' refers to an architecture with layer normalization before each sub-layer and transparent attention. Best results for each layer depth are shown in bold.}
\label{tab:deep_low}
\end{table}
\normalsize

\subsubsection{High resource}

In light of aforementioned experiments, we no longer train models with post layer normalization and restrict ourselves to pre layer normalization and transparent attention. Here, we also experiment with the {BPE} \numprint{30000} vocabulary. The results are shown in Table \ref{tab:deep_high}. Here again we see an improvement in BLEU scores with increasing depth of 1-2 points when going beyond 6 encoder layers. Transparent attention seems to help consistently for character level models but barely does anything for the two {BPE} models. Further, with increased depth, {BPE} \numprint{5000} and \numprint{30000} perform similarly in contrast to shallow models where there is a 1 BLEU difference. However, character level models are still slightly worse than the {BPE} models with a max score of 37.7 rather than 38 for the {BPE} models.

\small\addtolength{\tabcolsep}{-1pt}
\begin{table}[!htb]
\centering
\begin{tabular}{c|cc|cc|cc} & \multicolumn{2}{c|}{Char}      & \multicolumn{2}{c|}{\numprint{5000}}   & \multicolumn{2}{c}{\numprint{30000}}    \\
enc & PreN & PreN + T  & PreN & PreN+T & PreN & PreN+T \\ \hline
6   & 36.3       & 36.5              & 36.2     & 36.4              & \textbf{37.2}     & 36.9              \\
12  & 36.8       & \textbf{37.5}     & 37.1     & 36.9              & 37.4              & \textbf{37.5}     \\
18  & 37.3       & 37.4              & 37.6     & 37.7              & \textbf{37.9}     & 37.8              \\
24  & 37.7       & 37.7        & 37.6     & 37.6              & \textbf{37.8}     & \textbf{37.8}              \\
32  & 37.2       & 37.4              & 37.9     & \textbf{38} & {\ul \textbf{38}} & 37.9             
\end{tabular}
\caption{Results for the high resource setting.}
\label{tab:deep_high}
\end{table}
\normalsize

\section{Discussion and recommendations}

\textbf{Vocabulary Size.} We observed that in the low resource setting, BPE vocab size can be a very important parameter to tune, having a large impact on BLEU. However, the effect vanishes for the high resource setting, where performance is similar for a large range of vocabulary sizes. Character level models also tend to be competitive with BPE.

\noindent
\textbf{Lexicographical noise.} When trained on clean data, character-based models are more robust to natural and synthetic lexicographical noise than BPE-based models (these results confirm a trend already observed in \cite{BelinkovNoise}), however the trend fades away when similar kind of noise is introduced in the training data as well. Surprisingly, we observed that noise on training data might be acting as a regularizer for the large BPE vocabularies (breaking up large tokens into smaller ones) and improves results on clean inputs.

\noindent
\textbf{Domain shift.} Better results on 4 datasets over 5 with character based models for DE-EN;  character level models outperform BPE when evaluated on data sufficiently different from the training domain in  low resource; no significant differences for EN-DE. In DE-EN (high resource), character level models are  only better when testing on the WMT-Biomedical test set which has the largest proportion of unseen words.

\noindent
\textbf{Deep models.} In contrast to Cherry et al. (2018), we see a degradation of performance with increasing depth without any modification of the Transformer architecture. We can train deeper and more efficient character-based Transformers by switching the sequence of performing layer normalization. Doing so, we can  train models with up to 32 layers in the encoder. We are able to narrow but not close the gap between character and BPE. These tricks may also hold for other use cases where longer input sequences are needed (for instance: document level NMT).

\section{Conclusion}
In this work, we have studied the characteristics of different representation units in NMT including character-level models and BPE models with different vocabulary sizes. We observed that different representations can have very different behaviours with distinct advantages and disadvantages. In the future, we would like to investigate methods to combine different representations in order to get the best of all worlds.

%\pagebreak
\bibliographystyle{IEEEtran}
\bibliography{thesis-ref}

%%% Comment out the appendix
%\begin{comment}
\appendix

\section{Training Details}
\begin{table*}[!htb]
\centering
\begin{tabular}{lll}
Hyperparameter                            & Transformer Base & Our Version               \\ \hline
Encoder embedding dimension               & 512              & 512                       \\
Encoder fully forward embedding dimension & 2048             & 1024                      \\
Encoder layers                            & 6                & 6                         \\
Encoder attention heads                   & 8                & 4                         \\
Decoder embedding dimension               & 512              & 512                       \\
Decoder fully forward embedding dimension & 2048             & 1024                      \\
Decoder layers                            & 6                & 6                         \\
Decoder attention heads                   & 8                & 4                         \\
Share encoder-decoder embeddings          & X           & \checkmark
\end{tabular}
\caption{Hyperparameter Settings for the modified Transformer architecture.}
\label{tab:hyperparms}
\end{table*}

The training details for the low resource setting are as follows. Training is done on 4 GPUs with a max
batch size of 4000 tokens (per GPU). We train for
150 epochs, while saving a checkpoint after every epoch and average the 3
best checkpoints according to their perplexity on a
validation set. For each setting, we test all 6 combinations of
dropout in $[0.3, 0.4, 0.5]$ and learning rate in
$[5, 10] \times 10^{-4}$. Using the best dropout and learning rate combination, 5 models (with different random seeds) are trained.

The training details for the high resource setting are as follows. Training is done on 4 GPUs with a max
batch size of 3,500 tokens (per GPU). We train for
60 epochs, while saving a checkpoint after every epoch and average the 3
best checkpoints according to their perplexity on a
validation set. For each setting, we test all 3 combinations of
dropout in $[0.1, 0.2, 0.3]$ and set the max learning rate to be $5 \times 10^{-4}$. Due to the significantly larger computational requirements for this dataset, we only train one model for each configuration.

\section{Robustness to Noise}
\label{sect:robustnessNoise_full}

\begin{figure*}[h]
	\centering
	
    \begin{subfigure}[b]{.4\linewidth}
    \centering
		\includegraphics[width=\linewidth]{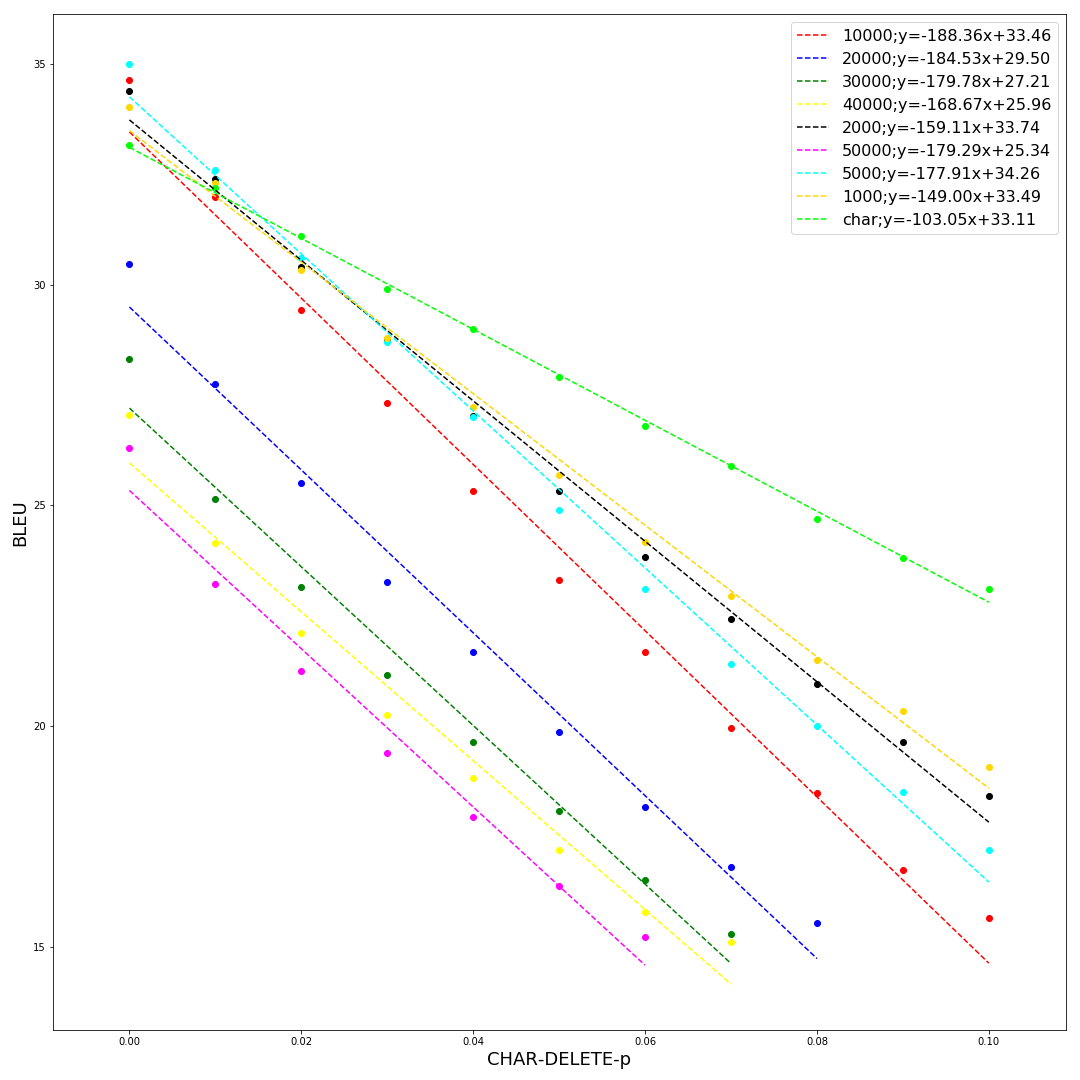}
	    \caption{\texttt{delete}}
	    \label{fig:delete-iwslt14-de-en}
	\end{subfigure}
	\begin{subfigure}[b]{.4\linewidth}
	    \centering
		\includegraphics[width=\linewidth]{./img/robustness/char-insert-iwslt14-de-en.png}
	    \caption{\texttt{insert}}
	    \label{fig:insert-iwslt14-de-en}
	\end{subfigure}
	
	\begin{subfigure}[b]{.4\linewidth}
    \centering
		\includegraphics[width=\linewidth]{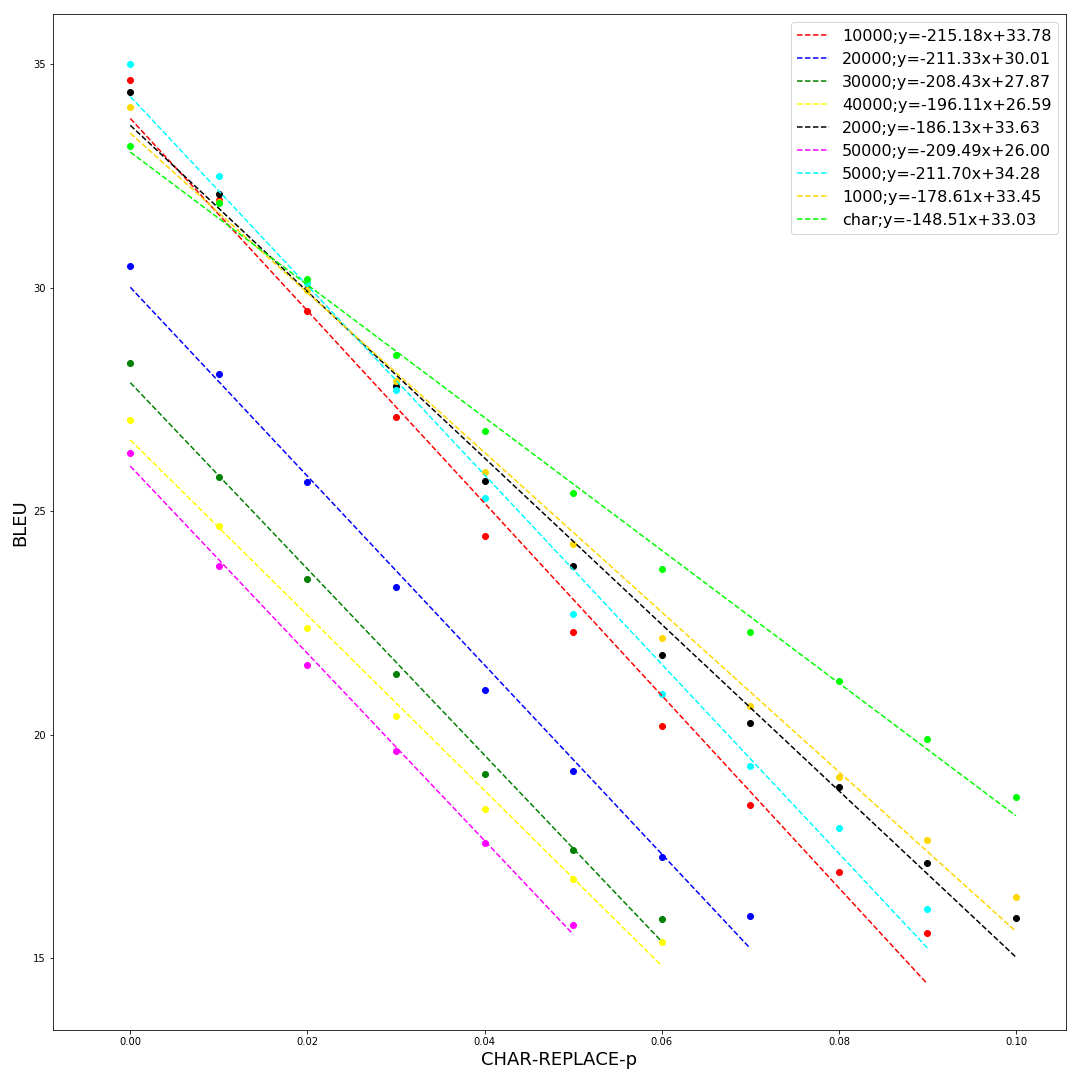}
	    \caption{\texttt{replace}}
	    \label{fig:replace-iwslt14-de-en}
	\end{subfigure}
	\begin{subfigure}[b]{.4\linewidth}
	    \centering
		\includegraphics[width=\linewidth]{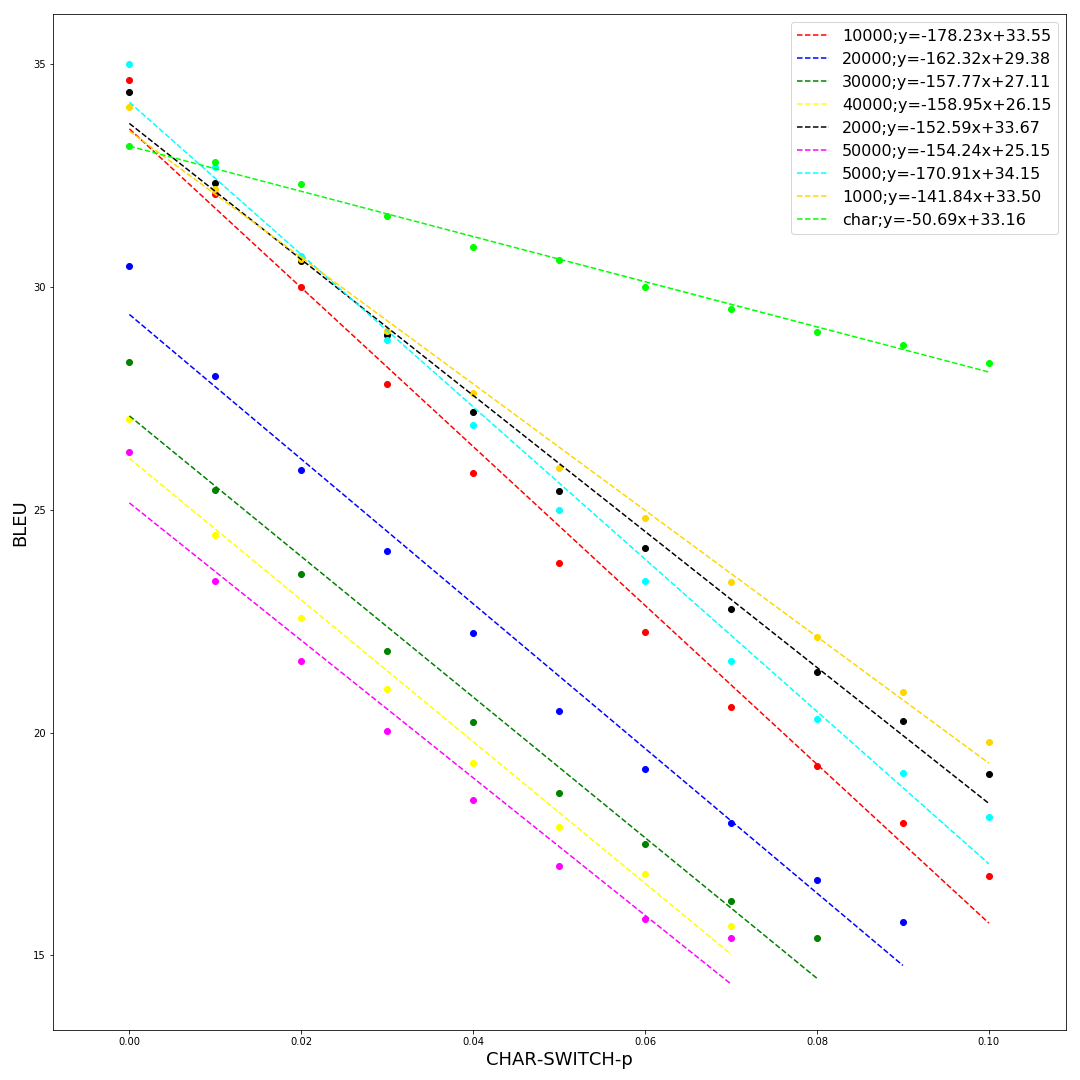}
	    \caption{\texttt{switch}}
	    \label{fig:switch-iwslt14-de-en}
	\end{subfigure}
	
	\begin{subfigure}[b]{.4\linewidth}
	    \centering
		\includegraphics[width=\linewidth]{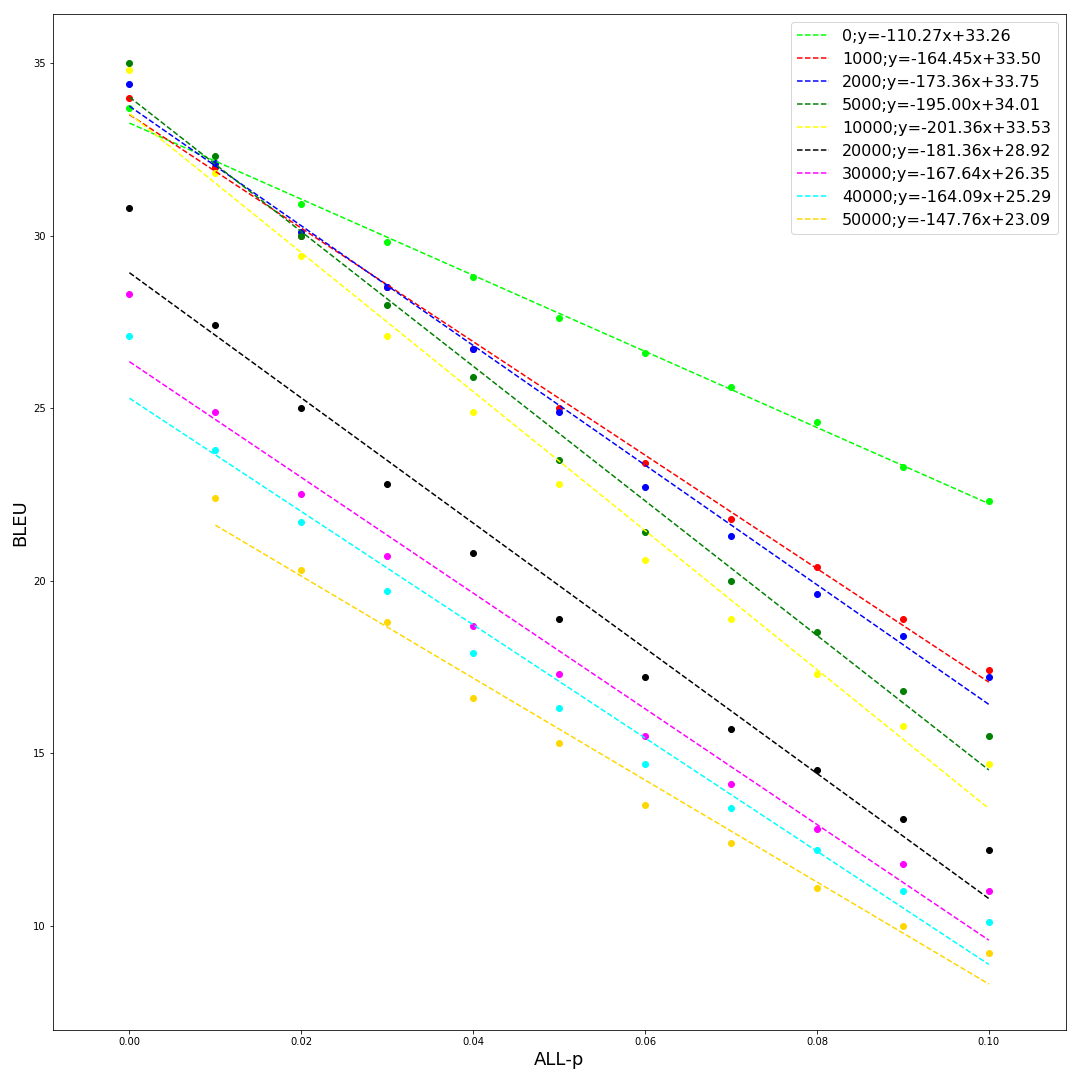}
	    \caption{\texttt{all}}
	    \label{fig:all-iwslt14-de-en}
	\end{subfigure}
	\begin{subfigure}[b]{.4\linewidth}
    \centering
		\includegraphics[width=\linewidth]{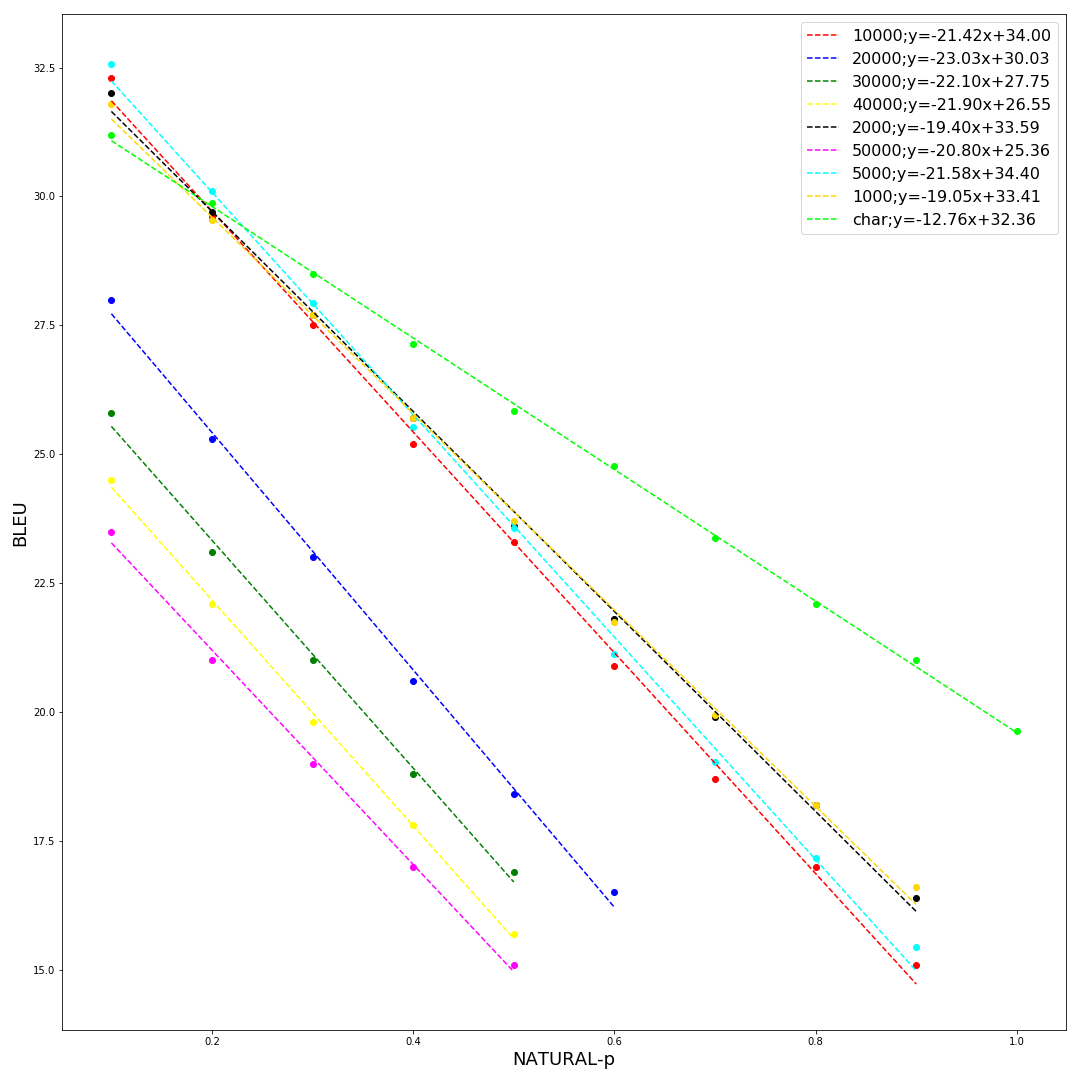}
	    \caption{\texttt{natural}}
	    \label{fig:natural-iwslt14-de-en}
	\end{subfigure}
	
	\caption{Degradation of translation quality with lexicographical noise for model trained on DE-EN language pair in the \textit{low resource} setting. Character level model is shown in light green (best viewed in color).}
	\label{fig:noise-impact-de-en-iwslt14}
\end{figure*}

\begin{figure*}[h]
	\centering
	
    \begin{subfigure}[b]{.4\linewidth}
    \centering
		\includegraphics[width=\linewidth]{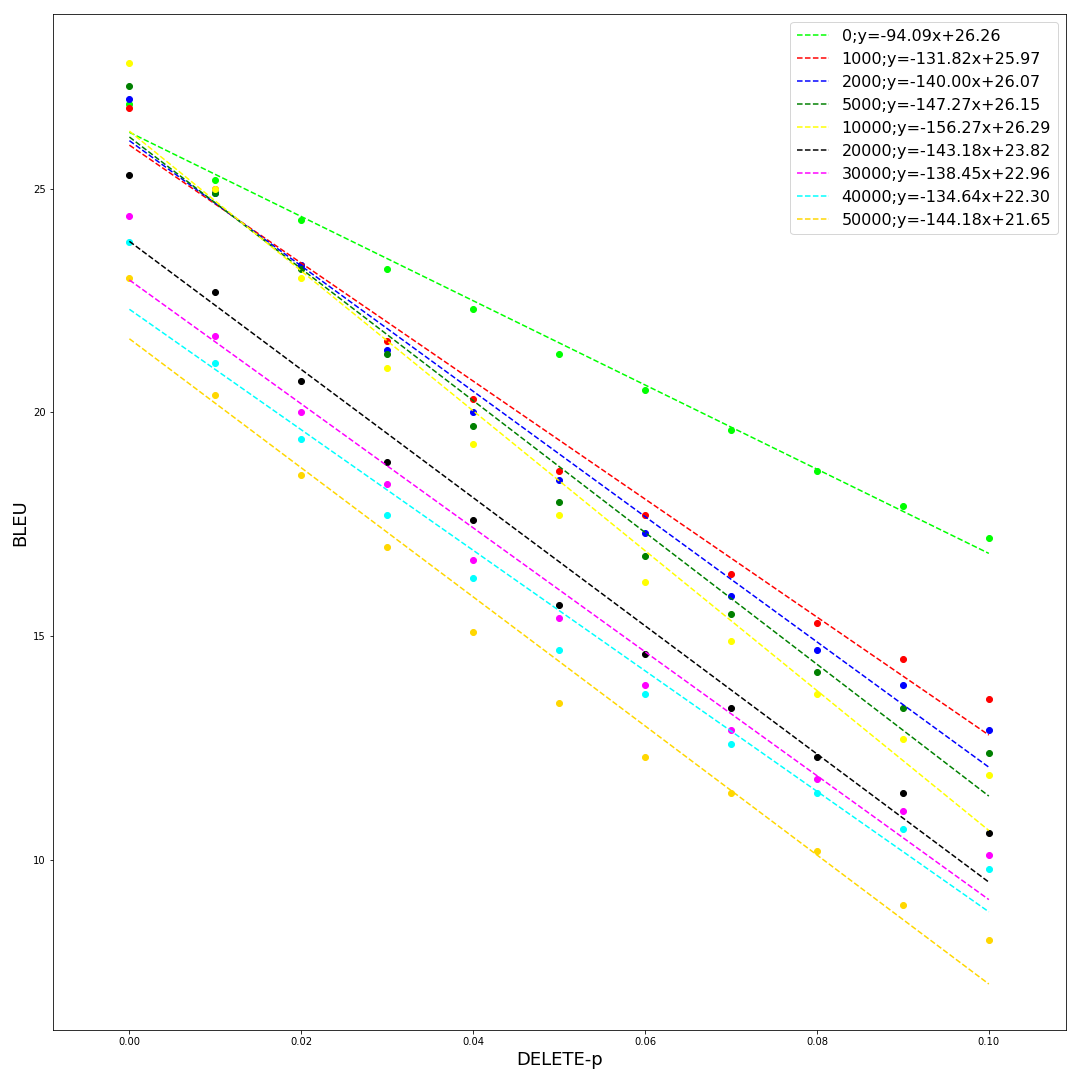}
	    \caption{\texttt{delete}}
	    \label{fig:delete-iwslt14-en-de}
	\end{subfigure}
	\begin{subfigure}[b]{.4\linewidth}
	    \centering
		\includegraphics[width=\linewidth]{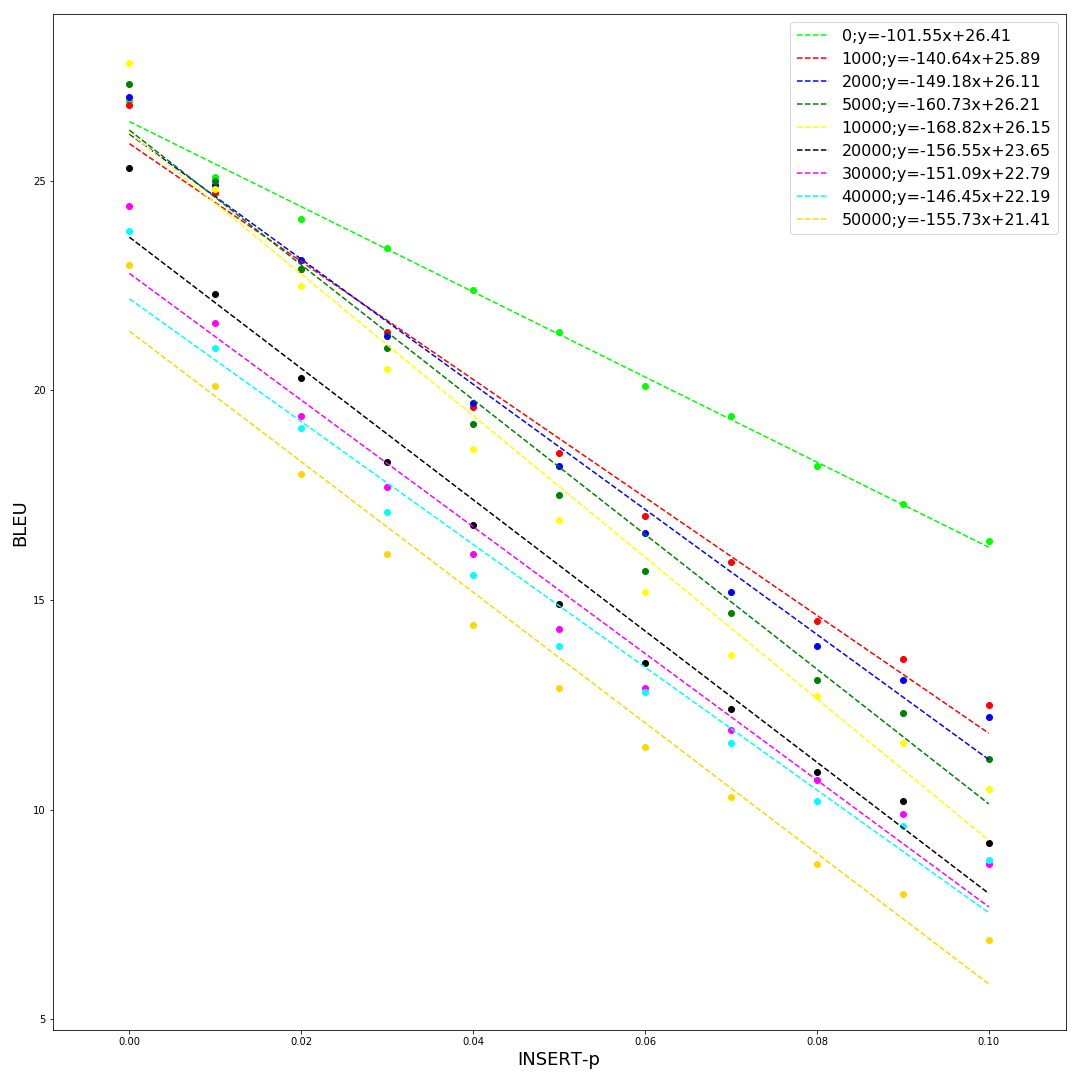}
	    \caption{\texttt{insert}}
	    \label{fig:insert-iwslt14-en-de}
	\end{subfigure}
	
	\begin{subfigure}[b]{.4\linewidth}
    \centering
		\includegraphics[width=\linewidth]{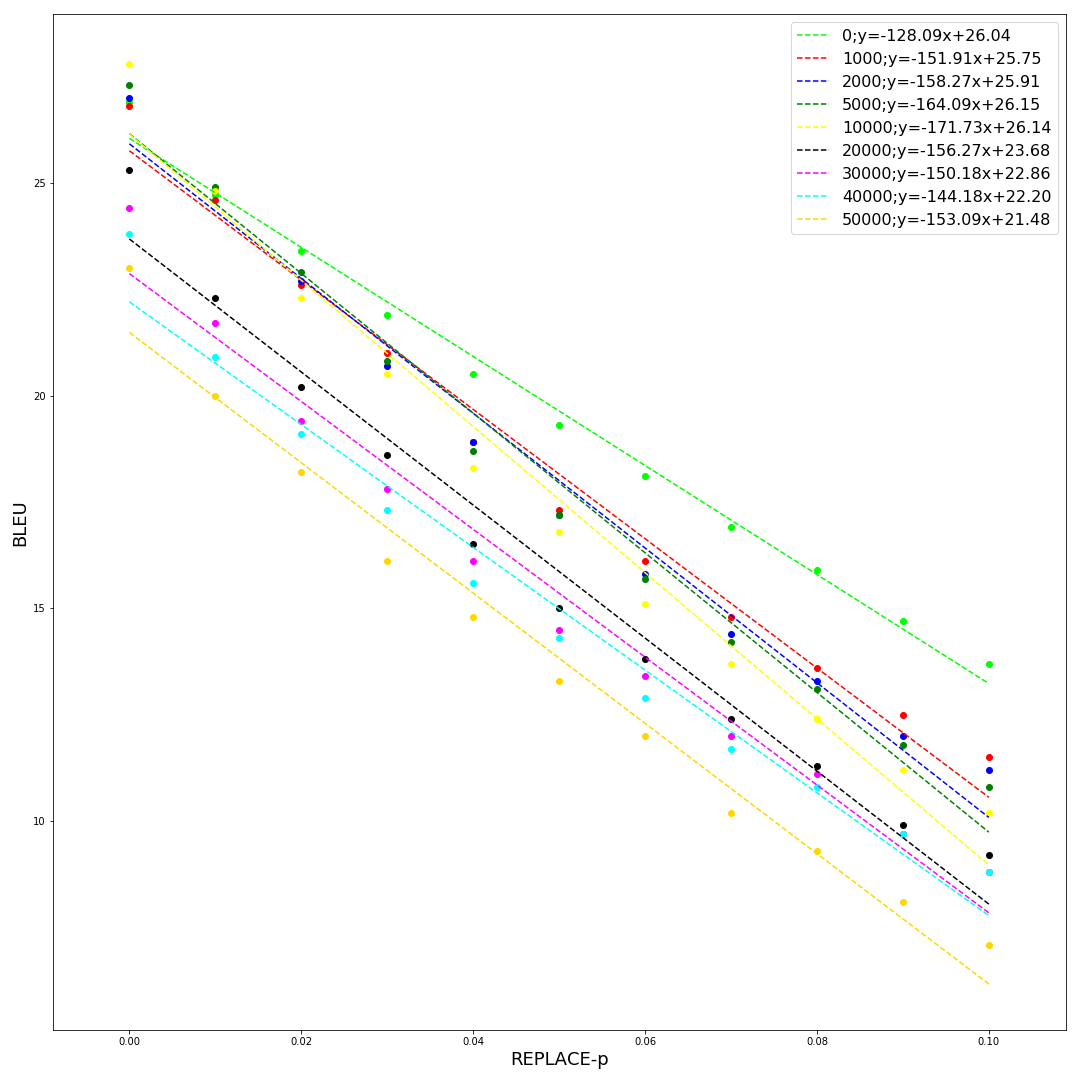}
	    \caption{\texttt{replace}}
	    \label{fig:replace-iwslt14-en-de}
	\end{subfigure}
	\begin{subfigure}[b]{.4\linewidth}
	    \centering
		\includegraphics[width=\linewidth]{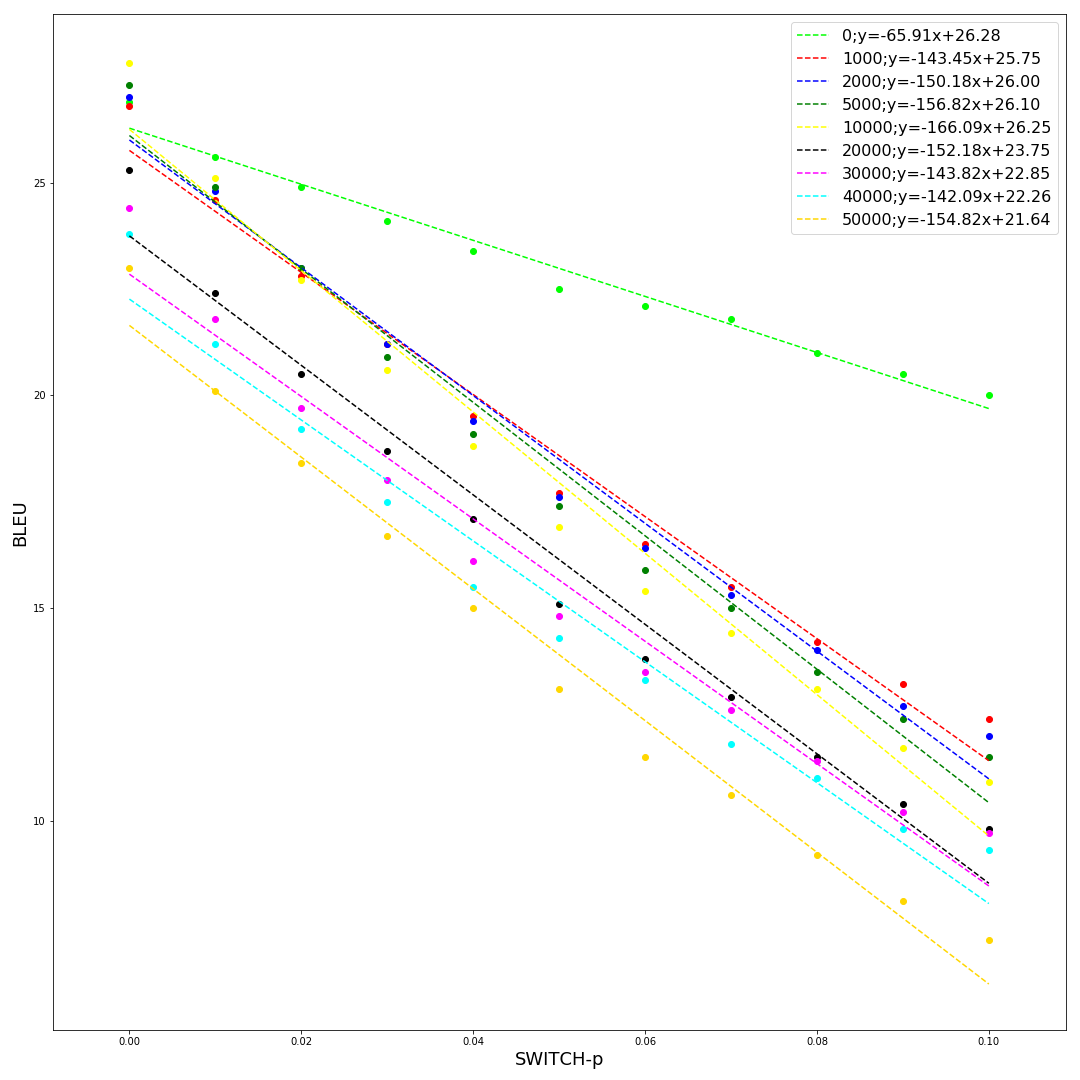}
	    \caption{\texttt{switch}}
	    \label{fig:switch-iwslt14-en-de}
	\end{subfigure}
	
	\begin{subfigure}[b]{.4\linewidth}
	    \centering
		\includegraphics[width=\linewidth]{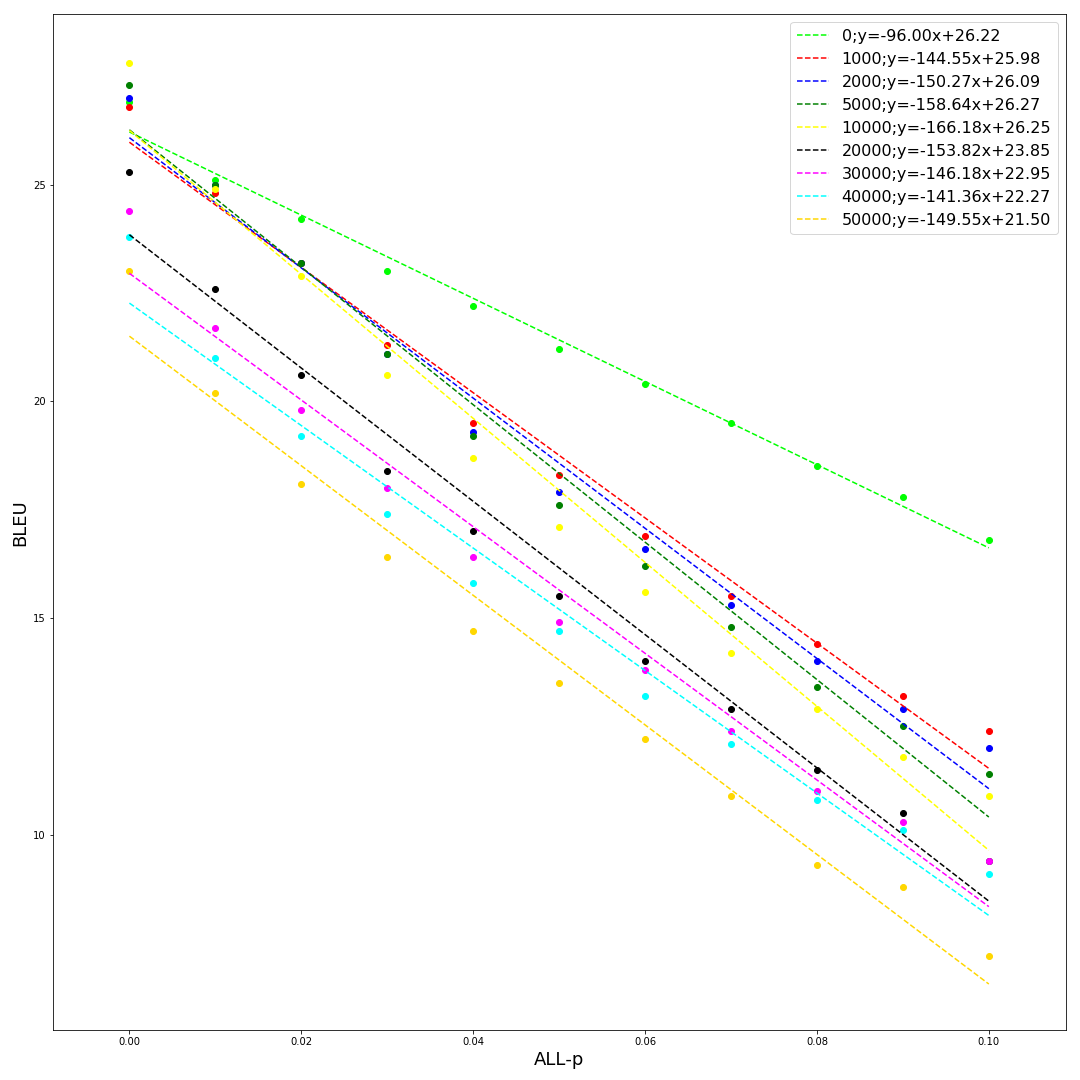}
	    \caption{\texttt{all}}
	    \label{fig:all-iwslt14-en-de}
	\end{subfigure}

	\caption{Degradation of translation quality with lexicographical noise for model trained on EN-DE language pair in the \textit{low resource} setting. Character level model is shown in light green (best viewed in color).}
	\label{fig:noise-impact-en-de-iwslt14}
\end{figure*}

\begin{figure*}[h]
	\centering
	
    \begin{subfigure}[b]{.4\linewidth}
    \centering
		\includegraphics[width=\linewidth]{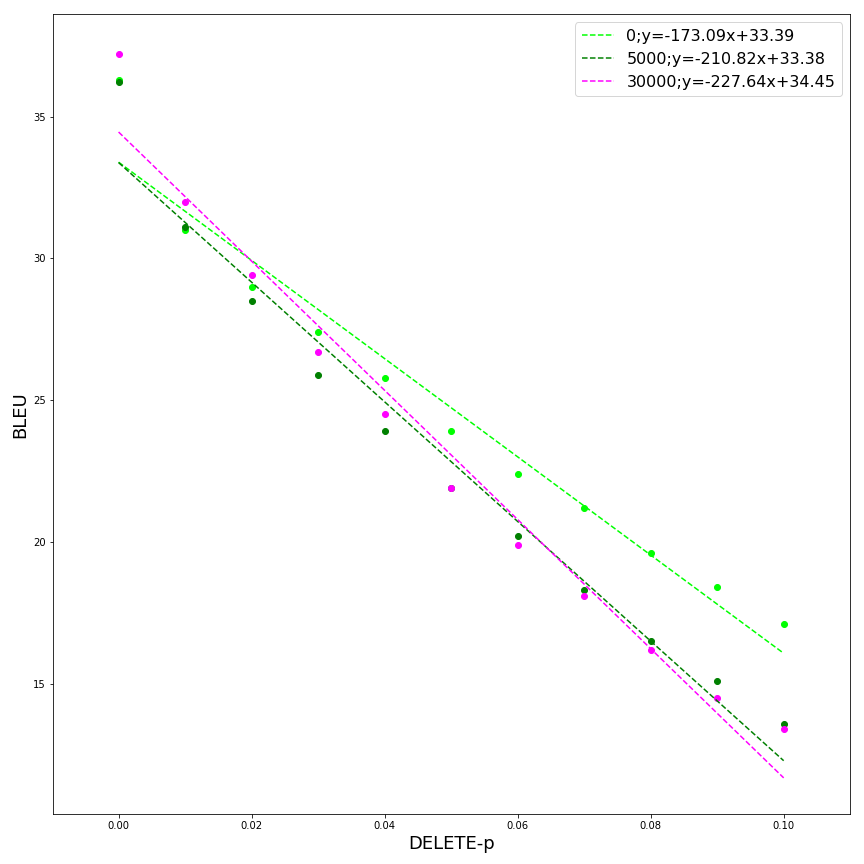}
	    \caption{\texttt{delete}}
	    \label{fig:delete-wmt-de-en}
	\end{subfigure}
	\begin{subfigure}[b]{.4\linewidth}
	    \centering
		\includegraphics[width=\linewidth]{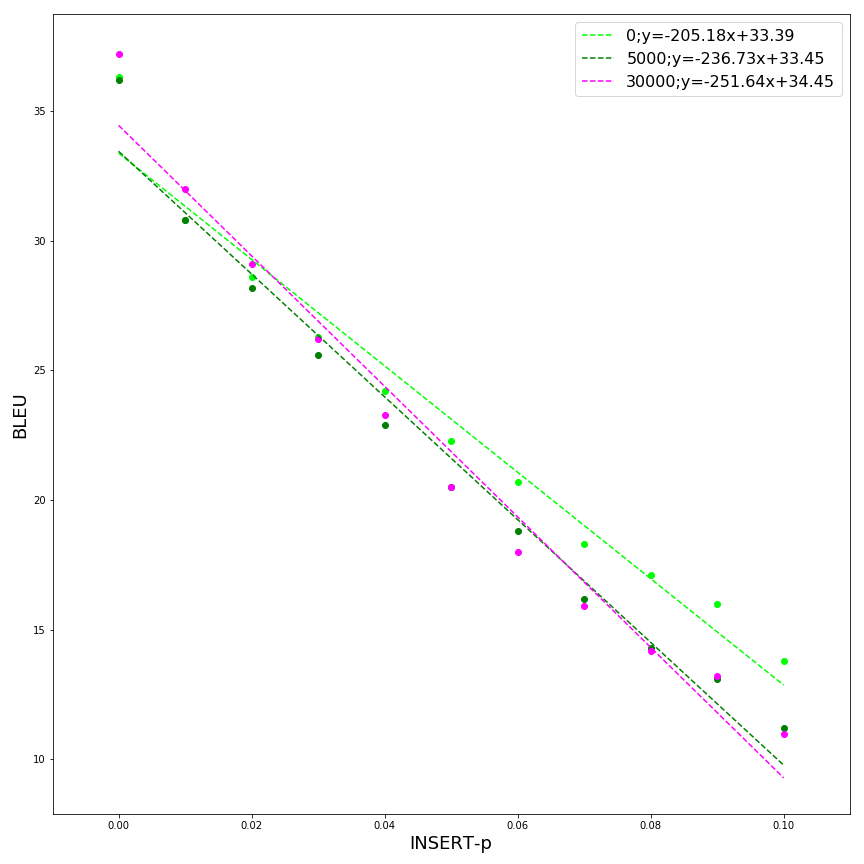}
	    \caption{\texttt{insert}}
	    \label{fig:insert-wmt-de-en}
	\end{subfigure}
	
	\begin{subfigure}[b]{.4\linewidth}
    \centering
		\includegraphics[width=\linewidth]{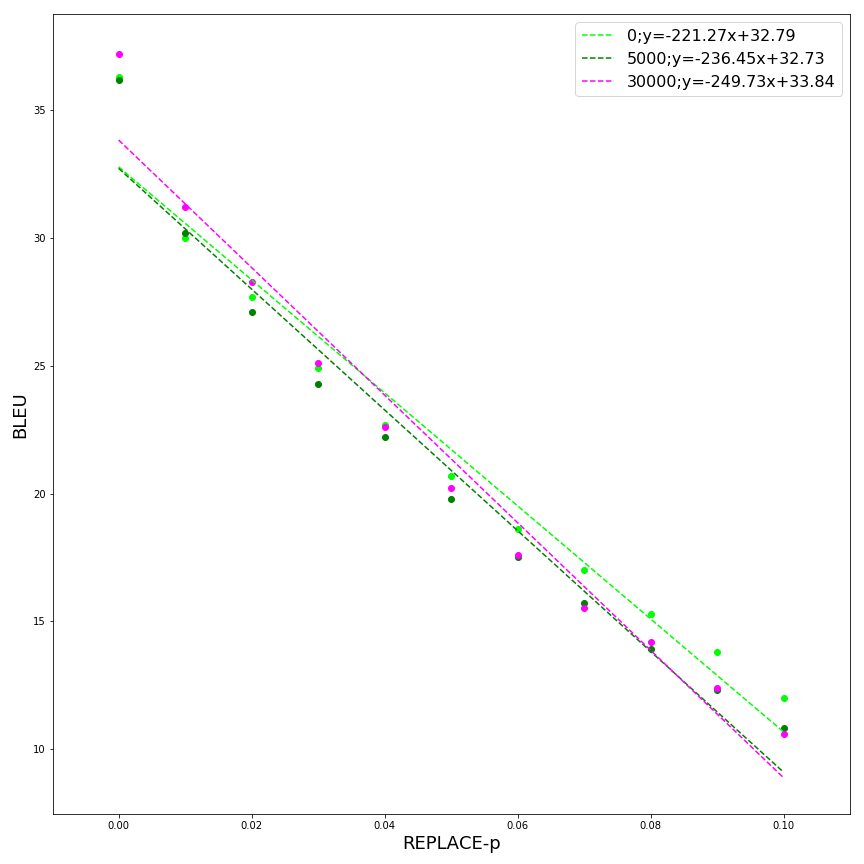}
	    \caption{\texttt{replace}}
	    \label{fig:replace-wmt-de-en}
	\end{subfigure}
	\begin{subfigure}[b]{.4\linewidth}
	    \centering
		\includegraphics[width=\linewidth]{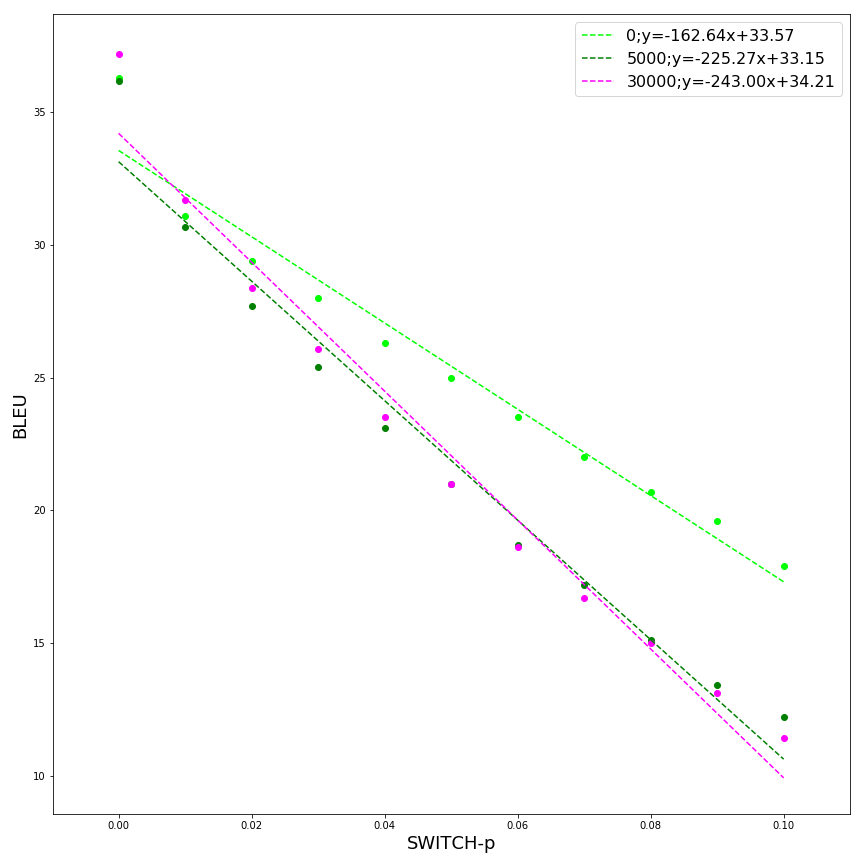}
	    \caption{\texttt{switch}}
	    \label{fig:switch-wmt-de-en}
	\end{subfigure}
	
	\begin{subfigure}[b]{.4\linewidth}
	    \centering
		\includegraphics[width=\linewidth]{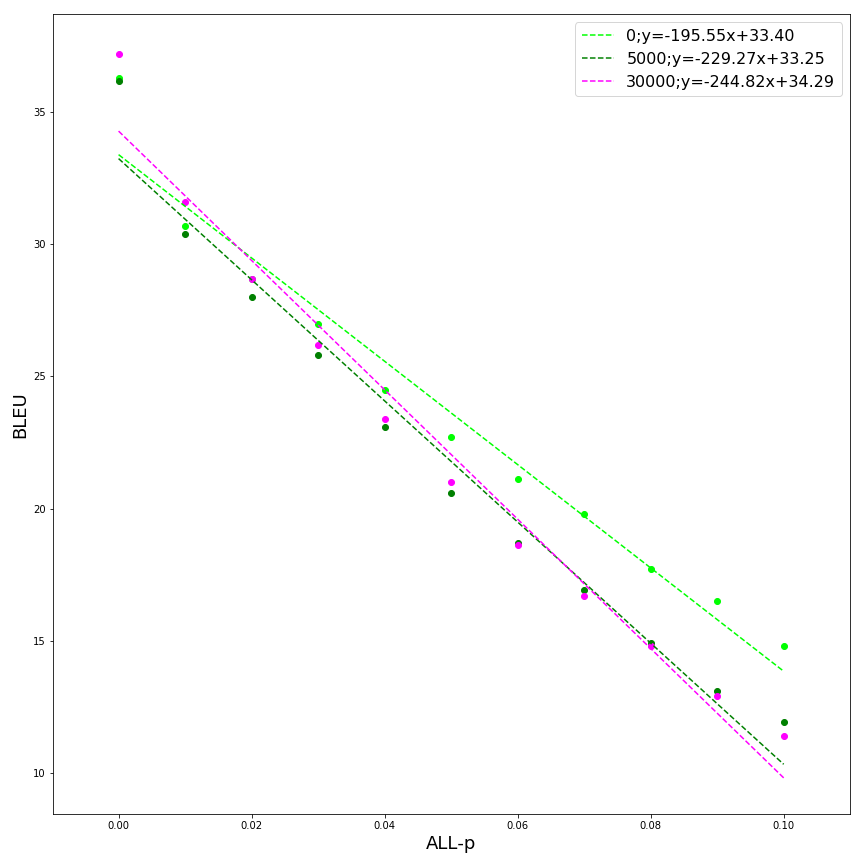}
	    \caption{\texttt{all}}
	    \label{fig:all-wmt-de-en}
	\end{subfigure}
	\begin{subfigure}[b]{.4\linewidth}
    \centering
		\includegraphics[width=\linewidth]{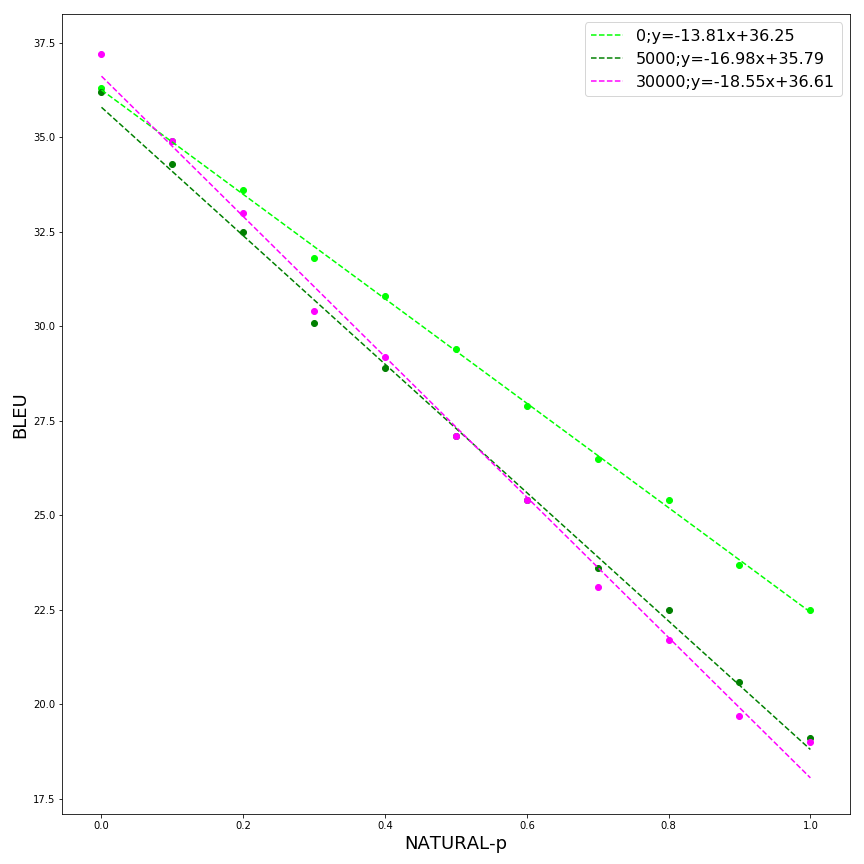}
	    \caption{\texttt{natural}}
	    \label{fig:natural-wmt-de-en}
	\end{subfigure}
	
	\caption{Degradation of translation quality with lexicographical noise for model trained on DE-EN language pair in the \textit{high resource} setting. Character level model is shown in light green (best viewed in color).}
	\label{fig:noise-impact-de-en-wmt}
\end{figure*}
%\end{comment}

%\pagebreak

%\appendix

We include here the full tables of training and testing on clean and noised datasets.
A summary of those tables can be found in Figure \ref{fig:noiseDEEN}  and \ref{fig:noiseENDE}
%Table~\ref{tab:train_noisy_data_de_en} and \ref{tab:train_noisy_data_en_de} 
in the main part of the paper.

\begin{table*}[h]
    \centering
    
\begin{subtable}{.9\textwidth}
\centering
\begin{tabular}{l|llllll|l}
train \textbackslash test     & clean         & \texttt{all}  & \texttt{delete} & \texttt{insert} & \texttt{replace} & \texttt{switch} & avg           \\ \hline
clean            & \textbf{34.1} & 27.6          & 27.9            & 27.7            & 25.4             & 30.6            & 28.9          \\
\texttt{all}     & 32.7          & \textbf{33.1} & 30.4            & 32.1            & 28.1             & 32.4            & \textbf{31.5} \\
\texttt{delete}  & 33.0          & 28.1          & \textbf{30.7}   & 25.9            & 24.6             & 30.6            & 28.8          \\
\texttt{insert}  & 32.7          & 28.8          & 25.5            & \textbf{32.4}   & 25.7             & 30.2            & 29.2          \\
\texttt{replace} & 32.9          & 29.7          & 28.2            & 30.3            & \textbf{31.1}    & 30.2            & 30.4          \\
\texttt{switch}  & 33.1          & 28.4          & 26.8            & 28.0            & 24.8             & \textbf{32.9}   & 29.0     
\end{tabular}
\subcaption{Character}
\label{tab:char_train_noisy_data_de_en}
\end{subtable}

\begin{subtable}{.9\textwidth}
\centering
\begin{tabular}{l|llllll|l}
train \textbackslash test     & clean         & \texttt{all}  & \texttt{delete} & \texttt{insert} & \texttt{replace} & \texttt{switch} & avg           \\ \hline
clean            & \textbf{35.0} & 23.5          & 24.9            & 23.5            & 22.7             & 25.0            & 25.8          \\
\texttt{all}     & 34.4          & \textbf{32.9} & 32.0            & 33.4            & 31.6             & 33.6            & \textbf{33.0} \\
\texttt{delete}  & 34.3          & 29.6          & \textbf{32.7}   & 27.3            & 25.5             & 31.1            & 30.1          \\
\texttt{insert}  & 34.4          & 30.9          & 29.1            & \textbf{33.8}   & 30.4             & 30.8            & 31.6          \\
\texttt{replace} & 34.0          & 30.7          & 29.2            & 32.1            & \textbf{31.9}    & 29.9            & 31.3          \\
\texttt{switch}  & 34.1          & 29.3          & 29.2            & 28.0            & 25.2             & \textbf{33.7}   & 29.9           
\end{tabular}
\subcaption{5,000}
\label{tab:5000_train_noisy_data_de_en}
\end{subtable}

\begin{subtable}{.9\textwidth}
\centering
\begin{tabular}{l|llllll|l}
train \textbackslash test     & clean         & \texttt{all}  & \texttt{delete} & \texttt{insert} & \texttt{replace} & \texttt{switch} & avg           \\ \hline
clean            & 28.2          & 17.3          & 18.2            & 17.0            & 17.9             & 19.0            & 19.6          \\
\texttt{all}     & 34.0          & \textbf{32.0} & 31.5            & 32.3            & 30.2             & 32.9            & \textbf{32.2} \\
\texttt{delete}  & \textbf{34.2} & 28.1          & \textbf{32.5}   & 25.5            & 24.2             & 29.3            & 29.0          \\
\texttt{insert}  & \textbf{34.2} & 30.4          & 28.8            & \textbf{32.9}   & 29.6             & 29.7            & 30.9          \\
\texttt{replace} & 33.8          & 30.4          & 28.8            & 31.5            & \textbf{30.9}    & 29.6            & 30.8          \\
\texttt{switch}  & \textbf{34.2} & 28.0          & 28.2            & 25.1            & 23.2             & \textbf{33.7}   & 28.7           
\end{tabular}
\subcaption{30,000}
\label{tab:30000_train_noisy_data_de_en}
\end{subtable}

    \caption{\textbf{BLEU} scores for DE-EN models trained and tested on different noises.}
    \label{tab:train_noisy_data_de_en_full}
\end{table*}

\begin{table*}[h]
    \centering
    
\begin{subtable}{.9\textwidth}
\centering
\begin{tabular}{l|llllll|l}
train \textbackslash test     & clean         & \texttt{all}  & \texttt{delete} & \texttt{insert} & \texttt{replace} & \texttt{switch} & avg   \\ \hline
clean            & \textbf{26.9} & 21.2          & 21.3            & 21.4            & 19.3             & 22.5            & 22.1               \\
\texttt{all}     & 25.7          & \textbf{25.3} & 24.1            & 26.2            & 24.2             & 25.5            & \textbf{25.2}      \\
\texttt{delete}  & 25.3          & 22.0          & \textbf{24.9}   & 20.2            & 19.8             & 23.2            & 22.6               \\
\texttt{insert}  & 25.9          & 22.8          & 20.7            & \textbf{26.5}   & 21.1             & 22.7            & 23.3               \\
\texttt{replace} & 26.0          & 23.5          & 21.9            & 24.3            & \textbf{25.3}    & 22.8            & 24.0               \\
\texttt{switch}  & 25.6          & 22.1          & 21.9            & 21.5            & 19.5             & \textbf{25.9}   & 22.8         
\end{tabular}   
\subcaption{Character}
\label{tab:char_train_noisy_data_en_de}
\end{subtable}

\begin{subtable}{.9\textwidth}
\centering
\begin{tabular}{l|llllll|l}
train \textbackslash test     & clean         & \texttt{all}  & \texttt{delete} & \texttt{insert} & \texttt{replace} & \texttt{switch} & avg           \\ \hline
clean            & \textbf{27.4} & 17.6          & 18.0            & 17.5            & 17.2             & 17.4            & 19.2          \\
\texttt{all}     & 25.9          & \textbf{25.4} & 24.7            & 26.0            & 24.9             & 25.5            & \textbf{25.4} \\
\texttt{delete}  & 26.1          & 22.1          & \textbf{25.5}   & 20.1            & 19.8             & 22.9            & 22.8          \\
\texttt{insert}  & 26.2          & 23.7          & 21.9            & \textbf{26.6}   & 23.4             & 22.6            & 24.1          \\
\texttt{replace} & 26.0          & 24.1          & 22.1            & 25.5            & \textbf{25.4}    & 22.5            & 24.3          \\
\texttt{switch}  & 26.3          & 22.4          & 22.6            & 20.9            & 19.9             & \textbf{26.2}   & 23.1         
\end{tabular}
\subcaption{5,000}
\label{tab:5000_train_noisy_data_en_de}
\end{subtable}

\begin{subtable}{.9\textwidth}
\centering
\begin{tabular}{l|llllll|l}
train \textbackslash test     & clean         & \texttt{all}  & \texttt{delete} & \texttt{insert} & \texttt{replace} & \texttt{switch} & avg           \\ \hline
clean            & 24.5          & 14.9          & 15.4            & 14.3            & 14.5             & 14.8            & 16.4          \\
\texttt{all}     & 26.2          & \textbf{25.2} & 25.1            & 25.7            & 24.6             & 25.7            & \textbf{25.4} \\
\texttt{delete}  & 26.2          & 21.6          & \textbf{25.6}   & 19.7            & 19.3             & 22.1            & 22.4          \\
\texttt{insert}  & 26.3          & 23.8          & 21.9            & \textbf{26.2}   & 23.7             & 22.4            & 24.1          \\
\texttt{replace} & 26.2          & 23.8          & 22.1            & 24.9            & \textbf{25.2}    & 22.4            & 24.1          \\
\texttt{switch}  & \textbf{26.6} & 21.7          & 21.5            & 19.5            & 19.0             & \textbf{26.4}   & 22.5         
\end{tabular}
\subcaption{30,000}
\label{tab:30000_train_noisy_data_en_de}
\end{subtable}

    \caption{\textbf{BLEU} scores for EN-DE models trained and tested on different noises.}
    \label{tab:train_noisy_data_en_de_full}
\end{table*}

\end{document}